\documentclass[lettersize,journal]{IEEEtran}
\usepackage{amsmath,amsfonts}
\usepackage{algorithmic}
\usepackage{algorithm}
\usepackage{caption}
\usepackage{array}
\usepackage[caption=false,font=normalsize,labelfont=sf,textfont=sf]{subfig}
\usepackage{textcomp}
\usepackage{picins}
\usepackage{stfloats}
\usepackage{url}
\usepackage{bbding}
\usepackage{multirow,booktabs}
\usepackage{verbatim}
\usepackage{graphicx}
\usepackage{cite}
\usepackage{soul} 
\usepackage{color, xcolor}

\begin{document}

\title{SelfOdom: Self-supervised Egomotion and Depth Learning via Bi-directional Coarse-to-Fine Scale Recovery}

\author{Hao Qu, Lilian Zhang*, Xiaoping Hu, Xiaofeng He, Xianfei Pan, Changhao Chen*
\thanks{The authors are with the College of Intelligence Science and Technology, National University of Defense Technology, Changsha, 410073, China}
\thanks{*Corresponding author: Changhao Chen (changhao.chen66@outlook.com) and Lilian Zhang (lilianzhang@nudt.edu.cn).}
\thanks{This work was supported by National Natural Science Foundation of China (NFSC) under the Grant Number of 62103427, 62073331, 62103430.}}

\markboth{In Submission,~Vol.~X, No.~X, Aug~2023}%
{Shell \MakeLowercase{\textit{et al.}}: A Sample Article Using IEEEtran.cls for IEEE Journals}

\maketitle

\begin{abstract}
Accurately perceiving location and scene is crucial for autonomous driving and mobile robots. Recent advances in deep learning have made it possible to learn egomotion and depth from monocular images in a self-supervised manner, without requiring highly precise labels to train the networks. However, monocular vision methods suffer from a limitation known as scale-ambiguity, which restricts their application when absolute-scale is necessary. To address this, we propose SelfOdom, a self-supervised dual-network framework that can robustly and consistently learn and generate pose and depth estimates in global scale from monocular images. In particular, we introduce a novel coarse-to-fine training strategy that enables the metric scale to be recovered in a two-stage process. Furthermore, SelfOdom is flexible and can incorporate inertial data with images, which improves its robustness in challenging scenarios, using an attention-based fusion module. Our model excels in both normal and challenging lighting conditions, including difficult night scenes. Extensive experiments on public datasets have demonstrated that SelfOdom outperforms representative traditional and learning-based VO and VIO models.
\end{abstract}

\begin{IEEEkeywords}
Visual Odometry, Depth Learning, Visual-inertial Odometry, Deep Neural Netework
\end{IEEEkeywords}

\section{Introduction}
\IEEEPARstart{L}{ocalization} is a crucial component in autonomous systems, as it provides accurate pose information necessary for path planning and decision-making in self-driving vehicles and mobile robots. Visual odometry (VO) is a popular positioning method that tracks and matches visual features between consecutive images to establish a multi-view geometry model for calculating relative pose. However, the estimates of pose from monocular VO are scale-ambiguous. Stereo VO, on the other hand, exploits a stereo camera to estimate camera pose with absolute scale and is generally more robust due to the fusion of information from two cameras. Visual-inertial odometry (VIO) is another approach that combines a camera with an inertial sensor to estimate motion states using filtering or nonlinear optimization. VIO can recover the global-scale and improve the accuracy and robustness of odometry estimation, particularly under complex lighting conditions.

Recently, deep learning has been used to develop end-to-end neural network models that can learn pose directly from images. These methods can be classified into two categories: supervised learning-based and unsupervised learning based approaches. Supervised learning based VO/VIO models require ground-truth pose labels to train the neural network \cite{deepVO,clark2017vinet}. Although these models perform well in challenging scenes, such as featureless areas or complex lighting conditions, labeling data for training the networks is expensive and time-consuming. Additionally, the trained models are difficult to generalize to new environments with different scene geometry and appearance. To address this, there have been several attempts at developing self-supervised learning-based VO/VIO models that can jointly learn to produce relative pose and scene depths by constructing a photometric loss between the real image and a wrapped image from novel view synthesis \cite{SfMLearner}.

Despite its advantages, self-supervised learning-based odometry estimation still suffers from the scale-ambiguity problem. To address this, researchers have attempted to incorporate prior information such as velocity \cite{packnet-sfm}, camera height \cite{SelfSupervisedSR} or pointcloud data \cite{VIOLearner} to recover the scale metric of pose and depth. VIOLearner \cite{VIOLearner} is one such method that uses point cloud data for scale recovery. By calculating the interpolated depths from point cloud data and using depths with absolute scale, VIOLearner can obtain pose with absolute scale. However, the quality of point cloud data greatly affects the performance of scale recovery as it is highly dependent on its sparsity. Furthermore, the interpolated depths do not conform to the multi-view geometry and cannot be used as supervised labels for depth learning. As a result, VIOLearner can only produce pose estimates. In contrast, our proposed method can generate both pose and depth estimates, overcoming these limitations.

We instead introduce \textbf{SelfOdom}, a novel \textbf{Self}-supervised \textbf{Odom}etry learning framework. This framework jointly produces pose and depth estimates with global scale metric from monocular image sequences. Our approach utilizes a bi-directional coarse-to-fine scale recovery strategy to achieve both visual-odometry (VO) and visual-inertial odometry (VIO) learning. Unlike other methods that directly interpolate depths, we utilize point clouds to compare against a pretrained depth model to calculate the coarse scale ratio of depths. Our proposed approach involves a two-stage process for scale recovery. Firstly, we produce coarse pose via single-directional scale recovery conditioned on coarse depths. Then, we refine depths and poses together through a bi-directional coarse-to-fine scale recovery. This strategy allows our model to accurately and robustly generate depths and poses with global scale under both normal and complex lighting conditions, such as a night driving scene. Furthermore, our framework is flexible and can accommodate inertial data in addition to images via an attention-based fusion module that further enhances the performance of rotation estimation.

Our contributions can be summarized as follows:

\begin{itemize}
   \item We propose \textbf{SelfOdom}, a novel self-supervised learning-based framework for estimating visual odometry or visual-inertial odometry. Our approach can jointly learn and produce poses and depths in absolute scale metric from monocular images. Moreover, SelfOdom is flexible enough to incorporate other sensors, such as IMUs, with images to further enhance prediction performance.
    \item A novel bi-directional coarse-to-fine learning strategy is introduced to the scale recovery process. We estimate the metric scale by comparing scaled depth predictions against a pretrained model, ensuring the consistency of photometric loss, followed by a coarse-to-fine training process to jointly refine pose and depth estimates.
    \item We conducted extensive experiments on public datasets and performed ablation studies to validate the effectiveness of our SelfOdom framework. Our approach demonstrates robust performance in predicting poses and depths under challenging conditions, such as complex lighting scenarios (driving scenes at night). This task is especially challenging for traditional VO/SLAM methods.
\end{itemize}

\section{Related Works}
\subsection{Traditional Visual Odometry}

\begin{figure*}[ht]
    \centering
    \includegraphics[width=14cm]{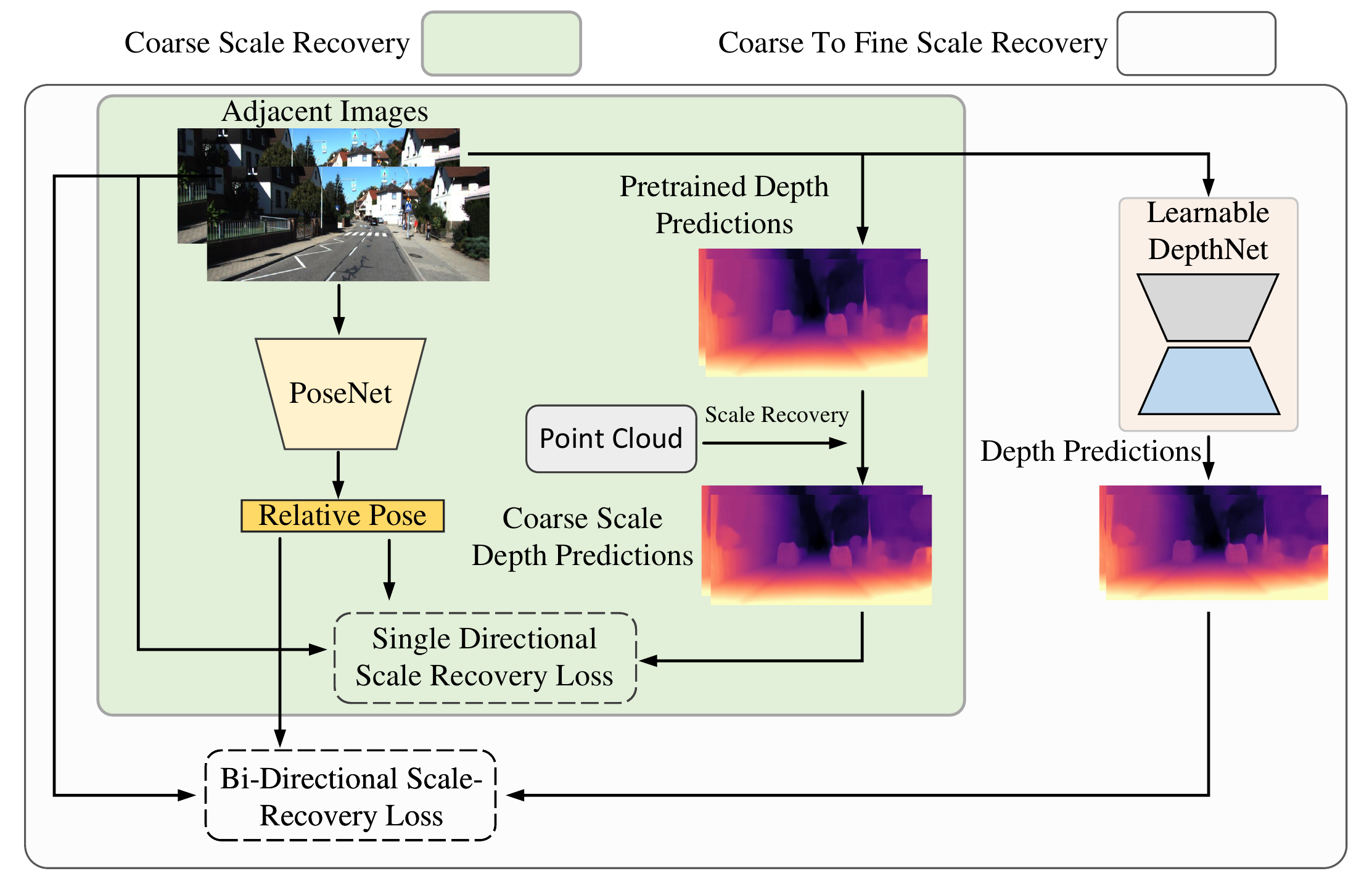}
    \caption{An overview of our proposed SelfOdom framework with a novel coarse-to-fine scale recovery process}
    \label{fig1:overview of our framework}
\end{figure*}

Visual odometry, or egomotion estimation, is a technique used to calculate the relative pose of a moving camera by matching feature points across different image frames using a multi-view geometry model. Several approaches have been developed in the literature to solve this problem. For instance, LIBVISO \cite{libviso} is a classic model-based visual odometry that uses two different types of convolution kernels to detect the maximum response points in the image as feature points. The RANSAC algorithm \cite{FISCHLER1987726} is then applied to filter the matching feature points. PTAM \cite{PTAM} uses the FAST algorithm \cite{FAST} to detect feature points on images with different resolutions, thus improving the performance of odometry at multi-scales. ORB-SLAM \cite{ORB} extends the application of visual odometry by adding stereo and depth cameras.
In environments where the texture is missing, direct-method shows better robustness in pose estimation than feature-point based methods. DSO \cite{DSO} and LSD-SLAM \cite{engel14eccv} use the corresponding pixel difference between adjacent images to construct a photometric loss model for calculating camera pose. However, under low-light environments, the performance of visual odometry decreases due to the failure of feature detection and matching. \cite{Venator2020RobustCP} adds semantic and GPS information in visual-based pose estimation framework. They conduct experiment in different driving datasets collected in the diverse environments. The result shows that their method still performs well under some complex conditions such as dynamic objects and few distinctive landmarks. Visual-inertial odometry (VIO) combines the measurements from a camera and an inertial measurement unit (IMU) to obtain better robustness and higher accuracy in pose estimation. For instance, \cite{forster2016manifold} uses IMU pre-integration to fuse the IMU raw data and optimizes the re-projection loss using matching feature points and point clouds. VINS \cite{VINS} adds automatic initialization, relocation, loop closure detection, and other functions to form a more complete visual-inertial odometry. Overall, these approaches have demonstrated promising results in egomotion estimation, enabling various applications in robotics and autonomous systems. In order to improve the performance of visual dead reckoning in urban areas, \cite{Takeyama2017ImprovementOD} introduces IMU information into the visual odometry to assist moving object detection. Besides, satellite doppler shift and IMU information are tightly coupled to correct the heading errors in the poor satellite environment. \cite{MultiModalFC} fuses visual and point cloud geometric features constraints in positioning and mapping framework. Their framework achieves robust performance in complex lighting condition. 
\subsection{The supervised learning of ego-estimation estimation}

Visual odometry (VO) has also seen advancements through the use of deep learning methods. DeepVO, a supervised learning-based VO, uses FlowNet's convolution layers to extract visual features, a Long Short-Term Memory (LSTM) network to integrate features from multiple images, and fully connected layers to transform features into pose predictions \cite{deepVO}. By utilizing accurate pose labels for training, DeepVO can learn to predict absolute pose predictions.
Clark et al. improved upon DeepVO by processing inertial data using an LSTM network to obtain motion features and combining them with visual features to form fused features \cite{clark2017vinet}. Selective attention mechanisms have also been introduced into visual-inertial odometry networks to improve robustness, as demonstrated in \cite{chen2019selective}. In \cite{chen2022learning}, point clouds are fused with inertial information to form a multi-sensor odometry estimation network.
Memory and refining modules have been proposed to preserve important contextual information and improve pose estimation, as demonstrated in \cite{xue2019beyond}. DAVO improves the performance of supervised learning-based VO by fusing inputs from semantic segmentation, optical flow, and RGB images via an attention module \cite{KuoLLLCL20}. DeepAVO, on the other hand, uses a channel-spatial attention-based PoseNet to learn frame-to-frame correspondence information from four branches of optical flow \cite{Zhu2021DeepAVOEP}.

\subsection{The Unsupervised learning of ego-estimation estimation}

Unsupervised learning of egomotion methods has gained popularity due to the absence of high-precision pose labels. SfMLearner \cite{SfMLearner} employs a double network structure, consisting of PoseNet and DepthNet, to estimate the relative pose between two consecutive images and predict the target frame depth, respectively. By combining relative pose estimates, target depth prediction, and target/source image pixels, the photometric loss is constructed.

\begin{figure*}[ht]
    \centering
    \includegraphics[width=14cm]{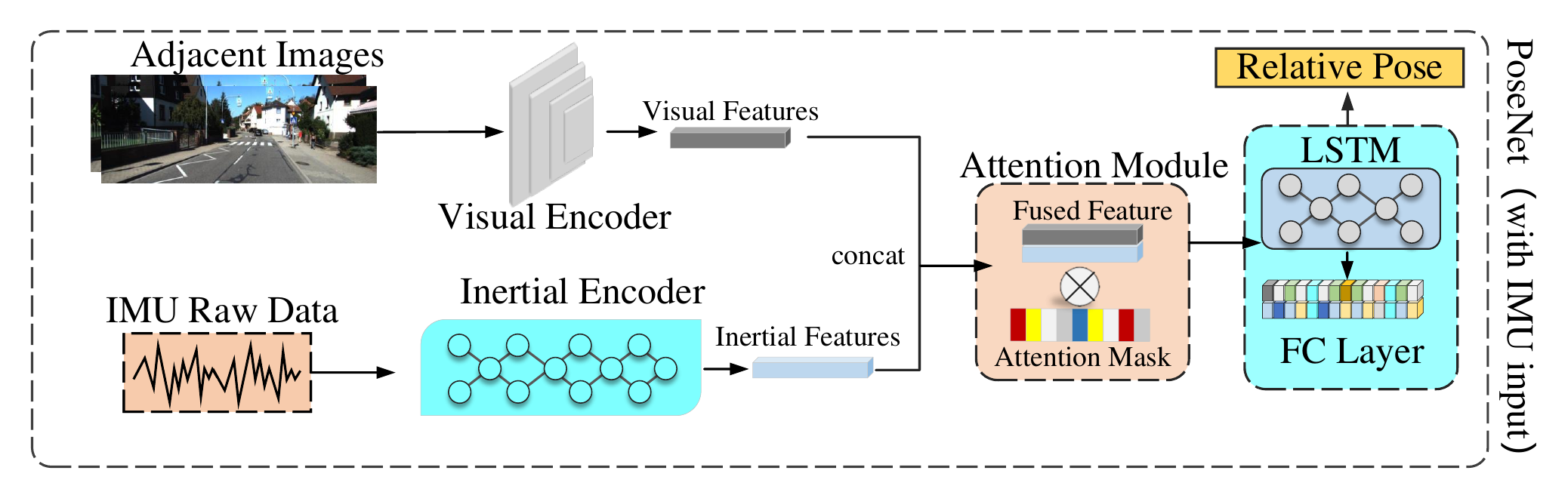}
    \caption{An overview of fusing visual and inertial information for pose estimation in our proposed SelfOdom. In visual odometry estimation, only visual information is used.}
    \label{fig: vio}
\end{figure*}

To enhance the accuracy of pose and depth estimates, Godard et al. \cite{godard2019digging} proposes a method that unifies the resolution of multi-scale depth estimates to the raw image resolution. Additionally, it uses the minimum value at each depth image pixel to form the photometric loss. In \cite{9710474}, disparity maps at all scales are predicted in the DepthNet encoder and decoder, and then transformed into pseudo-labels, which are utilized as self-distillation signals in the training process. To facilitate depth learning, depth hint labels are introduced into the photometric loss. Furthermore, a novel data augmentation method is proposed to avoid overfitting. Bian et al. \cite{bian2019unsupervised} addresses the issue of inconsistent depth prediction scales by proposing a depth scale consistency loss to constrain the depth and pose estimates to a unified scale. The depth prediction scale remains globally consistent throughout multiple training iterations. To distill geometric information in depth estimation, \cite{Liu2022SelfSupervisedMD} introduces geometric prior and pixel-level sensitivity in the whole framework. Their framework achieves state-of-art performance in both ego and depth estimation. PackNet-SfM \cite{packnet-sfm} introduces instantaneous velocity to calculate the adjacent frames' displacement as weak velocity supervision, which is combined with the relative displacement estimate from PoseNet to construct velocity supervision loss.
SelfSupervisedSR \cite{SelfSupervisedSR} estimates the height of the camera to the ground, which obtains the scale factor between the ground truth and the height estimate. This scale factor can recover the scale of the depth estimate and be introduced into the translation loss to accelerate the training convergence of the pose estimation. VIOlearner \cite{VIOLearner} incorporates IMU data into unsupervised work using a CNN to process IMU data and designs an online error correction module to calculate multi-scale photometric errors. The sparse point clouds are interpolated to ground truth depth labels, and then integrated into the photometric error with adjacent images. The odometry performance is better than that of end-to-end VINet, but this work does not include depth estimation, so the entire scene cannot be reconstructed. Self-VIO \cite{almalioglu2022selfvio} transforms IMU data into inertial features through CNN and combines them with visual features to obtain fusion features. 
The proposed approach incorporates an attention layer to filter fusion features and a GAN discriminator to optimize depth prediction for high-frequency structure. Additionally, an LSTM network is used to comprehensively optimize multi-frame pose and depth predictions. Experimental results indicate that the proposed self-VIO method performs well in complex environments. On the other hand, UnVIO\cite{UnVIO} introduces two optimization modules, namely intra-window and inter-window optimization. Intra-window optimization constructs photometric error between adjacent frames and a smooth loss function. In inter-window optimization, integrated information and pose prediction of odometry are used to construct trajectory consistency loss and 3D geometric consistency loss. The two training losses work to constrain the global pose scale and make pose prediction more consistent. However, it should be noted that the network is unable to obtain pose results with absolute scale.

\section{Self-supervised Visual Egomotion and Depth Learning}

In this section, we discuss the network architecture and the scale recovery method employed in the proposed \textbf{SelfOdom} framework. Figure {\ref{fig1:overview of our framework}} presents an overview diagram of the framework, which follows a double network design consisting of PoseNet and DepthNet, inspired by \cite{SfMLearner}. As depicted in Figure {\ref{fig: vio}}, the PoseNet comprises a visual encoder, an inertial encoder, an attention-based feature fusion module, and a pose estimation module. The framework takes a sequence of images as input and produces corresponding poses and depth maps. Initially, a pair of adjacent images undergoes multi-state visual feature extraction using a ResNet18-based visual feature encoder. The temporal dependency of these features is then modeled using a Long Short-Term Memory (LSTM) network, and the visual features after the LSTM are processed through fully connected layers to obtain the relative pose between the source and target frames. The DepthNet estimates the depth of each image in the sequence. In addition, an inertial feature extractor can be incorporated into the framework to create a visual-inertial PoseNet. The inertial features are encoded from a sequence of inertial measurement units (IMUs) and combined with the visual features using an attention module before being fed through the LSTM network.

The proposed framework is trained in two steps: (1) Coarse Scale Recovery, where the PoseNet output poses have a coarse absolute scale, and (2) Coarse-to-Fine Scale Recovery, where the PoseNet pose prediction is refined, and the DepthNet parameters are updated. The Coarse-to-Fine Scale Recovery step employs a coarse-to-fine scale recovery strategy to enhance the accuracy and robustness of the framework.

\subsection{ {Framework Modules}}
\subsubsection{Visual Encoder}
We use ResNet18 backbone as the visual feature encoder $f_\text{visual}$.  {Within a single sequence window, there are n consecutive images. The length of the sequence window is the total number of images in the sequence.
} Two adjacent images $\left \{\mathbf{I}_{i}, \mathbf{I}_{i+1}\right \} $ are combined on the RGB channel to obtain 6-dimensional tensors, which are inputted into the visual feature encoder to obtain visual features in 1/4 input image size. In order to simplify the network structure and improve the forward efficiency, we use global average pooling to transform ith frame visual features into a 128 dimensional feature vector $\mathbf{v}_i$. 

\begin{equation}
\label{eq: visual features}
\mathbf{v}_i= f_\text{visual}\left \{ \mathbf{I}_{i}, \mathbf{I}_{i+1} \right \} 
\end{equation}

\subsubsection{Inertial Encoder}
The data acquisition frequency of IMU is higher than that of camera, and it has high measurement accuracy in a short time. Between two adjacent images $\left \{\mathbf{I}_{i}, \mathbf{I}_{i+1}\right \} $, there exist $m$ imu raw data  {$\mathbf{M}_{i\to{i+1}}$}.  The jth imu frame includes linear acceleration $\alpha_{j}$ and angular velocity $\omega_j$ in body coordinate, is one-dimensional time series data, so we introduce the Bi-directional LSTM network as inertial feature encoder $f_\text{inertial}$, which excels at processing time series data. The input dimension of two-layer Bi-directional LSTM network is 6, the number of hidden state $\left \{\mathcal{H}_{j-1},\mathcal{H}_{j}  \right \} $is 128, and the dimension of jth frame output inertial feature vector $\mathbf{a}_i$ is 256.  {The range of subscript j is between [0,m-1]. And we use the output of $f_\text{inertial}$  at $j=m-1$ as the inertial feature $\mathbf{a}_i$ between two adjacent images.}

\begin{equation}
\label{eq: IMU Raw data}
 {\mathbf{M}_{i\to{i+1}  }=\begin{bmatrix}
  \alpha^{0} & \omega^{0}  \\
  \cdots &\cdots \\
  \alpha^{j}& \omega^{j} \\
  \cdots &\cdots \\
  \alpha^{m-1}& \omega^{m-1}
\end{bmatrix}\in \mathbb{R}^{m\times 6}}
\end{equation}

\begin{equation}
\label{eq: inertial features2}
\mathbf{a}_{j}, \mathcal{H}_{j} =f_\text{inertial}\left \{({\alpha^{j}},{\omega^{j}}  );\mathcal{H}_{j-1}    \right \}
\end{equation}

\begin{equation}
\label{eq: inertial features}
 {\mathbf{a}_{i}=\mathbf{a}_{m-1} }
\end{equation}

\subsubsection{Attention based Feature Fusion Module}
There may be a few misalignment in the raw data, bias in the IMU data, and errors in the calibration of intrinsics and extrinsics parameters. If the features of the two kind of data are directly combined, the odometry performance is likely to be suboptimal. Even in the pure visual feature, image features with missing texture will weaken the performance of pose estimation. Inspired by \cite{chen2019selective},  {we introduce the soft attention module in PoseNet as shown in Figure {\ref{fig: vio}}. The soft attention module can learn end-to-end and generate a weight mask with values ranging between (0,1), with the size of the mask being consistent with that of the concatenated features. Each value in the weight mask denotes the importance of the corresponding concatenated feature. Our rationale behind using the soft attention module is to ensure that the feature vectors remain free of any noise during the training process, which we believe can aid in achieving training convergence. Over several epochs of training, the weight mask assigns higher weights to the features that are crucial for achieving optimal training convergence.} We concatenate two kind of features $\left \{\mathbf{a_i}; \mathbf{v_i} \right \}$ to form the concatenated features.  {Subsequently, we employ a fully connected layer followed by a sigmoid layer to create the attention mask $\mathbf{s_i}$. We then re-weight each element of the concatenated features by multiplying it with the corresponding value of the soft attention mask\cite{chen2019selective}.} In the end, we obtain the filtered fusion features $\mathbf{z_i}$. 

\begin{equation}
\label{eq: masked fusion features}
 {\mathbf{s_i} = \text{Sigmoid} \left \{\text{FC} \left \{\mathbf{a_i}; \mathbf{v_i} \right \}  \right \} }
\end{equation}

\begin{equation}
\label{eq: filtered fusion features}
 {\mathbf{z_i} = \mathbf{s_i}\odot\left \{ \mathbf{a_i}; \mathbf{v_i} \right \}}  
\end{equation}

\subsubsection{Pose Estimation Module}
The filtered fusion features are brought into pose estimation module. In this module, there exist a LSTM network and a FC layer.  {The Long Short-Term Memory (LSTM) network integrates the multi-filtered fusion features $\mathbf{z_{i}}$. Within a single window, there are n-1 fusion features present since every pair of adjacent frames generates a fusion feature.} We set the hidden layer nodes in LSTM network to 1024 and the high-level output features $\mathbf{ \bar{z}_{i}}$ dimension to 2048. After the LSTM network, we use a dropout layer with a coefficient of 0.2 to further prevent the possibility of over-fitting. We subsequently send the high-level output features to the FC layer to obtain 6-DOF relative poses $\mathbf{T^{i+1}_{i}}$  {between adjacent two frames.}

\begin{equation}
\label{eq: Pose Estimation Module_lstm}
\mathbf{ \bar{z}_{i}} = \text{LSTM} \left \{\mathbf{{z}_{i}}\right \} _{i}
\end{equation}

\begin{equation}
\label{eq: Pose Estimation Module_FC}
\mathbf{T^{i+1}_{i}} = \text{FC} \left \{\mathbf{ \bar{z}_{i}}\right \} _{i}
\end{equation}

\begin{figure*}[ht]
    \centering
    \includegraphics[width=8cm]{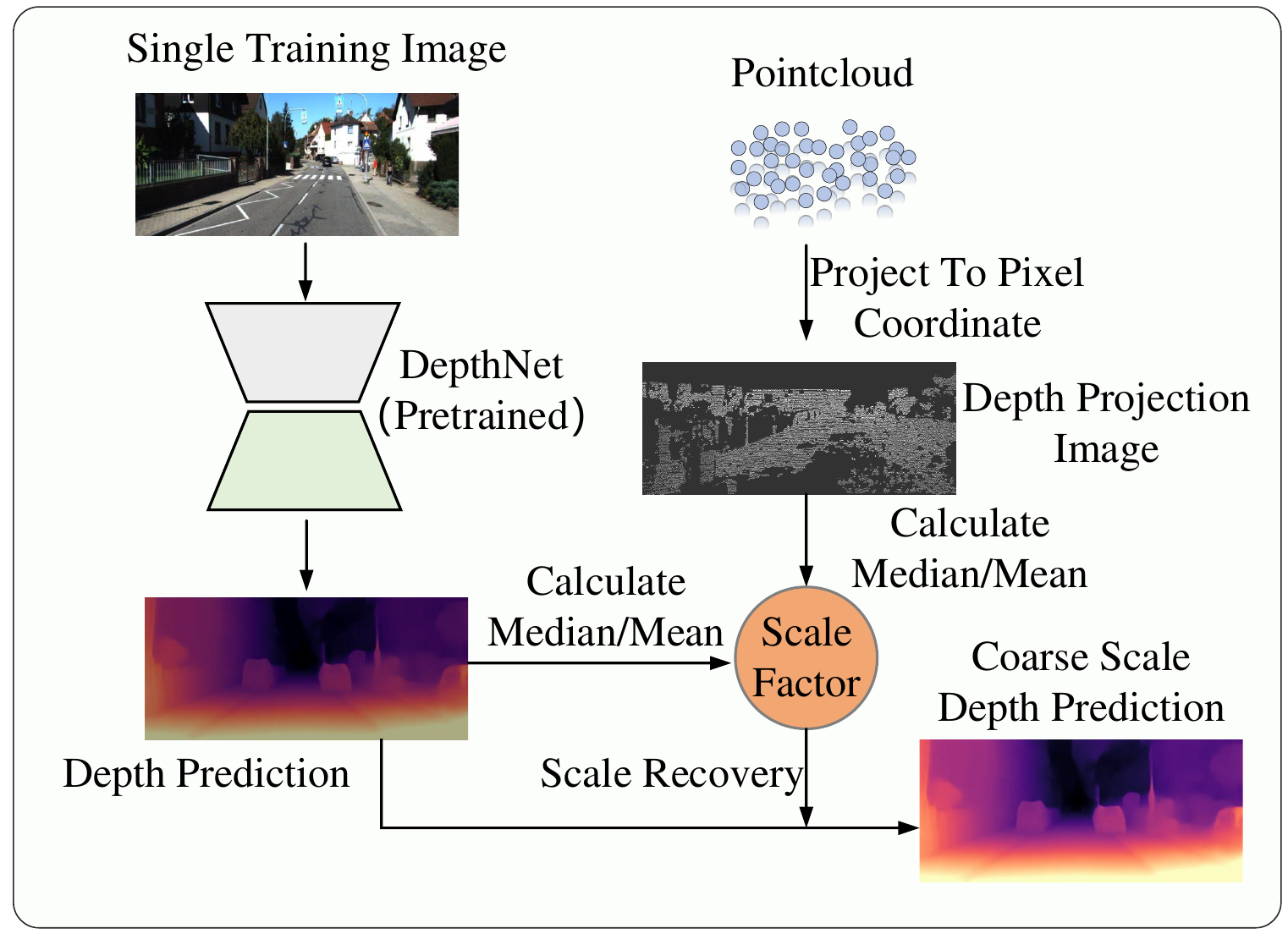}
    \caption{An overview on coarse depth scale recovery.}
    \label{fig2:Generation of depth labels}
\end{figure*}

\subsubsection{DepthNet} 
We use U-Net like DepthNet, which is divided into encoder and decoder. The encoder of the DepthNet adopts the ResNet18 backbone, which is pretrained on ImageNet data set. DepthNet encoder can extract the visual feature in input image. To improve the accuracy of low-level detail in depth predictions, the decoder component of the DepthNet combines feature maps from multi-scale encoder layers. Additionally, to upsample the visual features to the original image resolution, we have designed multiple convolution layers and incorporated nearest neighbor interpolations within the DepthNet decoder. We take the DepthNet decoder output as the disparity map. In the end we use the reciprocal of the disparity map as the depth prediction. 

\subsubsection{Photometric loss}
The pixel points of the same 3D projection in two adjacent frames have brightness constancy and spatial smoothness.
Given the estimated relative pose between the target frame $\mathbf{I}_t$ and the source frame $\mathbf{I}_s$, the depth of target frame $\mathbf{D}_{t}$ and camera intrinsic $\mathbf{K}$, we can generate the synthesized image $\mathbf{\hat{I}}_s$ by warping $\mathbf{I}_s$ into target-view:

\begin{equation}
\label{eq: photometric loss}
\mathbf{p}_s\sim \mathbf{K} \mathbf{T}_{t}^{s} \mathbf{D}_{t}(\mathbf{p}_t) \mathbf{K}^{-1} \mathbf{p}_{t},
\end{equation}
where $\mathbf{p}_s$ and $\mathbf{p}_t$ indicate pixel point in source frame and target frame. 

This work aims to recover the scale of ego-motion and depth learning. If the odometry pose scale is not recovered, it is difficult for odometry to be applied in realistic scenes. The reasons why the monocular unsupervised odometry cannot recover the pose scale: 1. The monocular image cannot determine the absolute distance of the 3D point in the single image. 2. It is impossible to determine the scale of $\mathbf{T}_{t}^{s}$'s translation vector and $\mathbf{D}_{t}$ only with photometric loss as a constraint.

\subsection{Coarse Depth Scale Recovery via Photometric-loss-consistent Pretrained Model} 
We first present a novel method for depth scale recovery through a photometric loss consistent pretrained model. Unlike previous approaches \cite{VIOLearner,VIOcompletion}, we do not employ dense depth completion results as supervisions. This is due to the fact that the performance of depth completion is closely related to the density of point clouds. In situations where point clouds are too sparse, the performance of dense depth completion may be degraded. Dense depth completion methods can be categorized into interpolation-based and learning-based methods. Interpolation-based methods complete sparse point clouds through interpolation, but the interpolated areas may not conform to multi-view geometry, resulting in unsuitable depths for learning depth recovery \cite{VIOLearner}. Learning-based methods, on the other hand, learn to generate dense depths from sparse depths, which can mitigate such issues to some extent, but there are still completion areas that do not conform to multi-view geometry, such as those that fall outside the LIDAR scanning range \cite{VIOcompletion,selfcompletion}. Additionally, it is challenging to remove outliers before using these completion-based depths as supervisions in photometric loss for scale recovery.

To address the aforementioned challenge, our framework proposes the use of a pre-trained DepthNet to generate depth predictions, and the scale of these predictions is recovered by comparing them with the corresponding point clouds. Unlike the previous methods that use dense depth completion, the depth predictions from the pre-trained model naturally conform to the multi-view geometry. Additionally, the depth predictions from the DepthNet are smoother and more complete, with no need to remove abnormal values in advance. We leverage the scale-recovered depth predictions as supervisions for the recovery of pose and depth estimates.

We firstly transform the point cloud vector $\mathbf{v}_{4\times1}$ from LIDAR to the pixel coordinate as the depth projection point $\mathbf{d}_{v}$. 

\begin{equation}
\label{deqn_D_vel}
\mathbf{d}_{v} = \mathbf{P}_{c2} \mathbf{T}^\text{rect}_{c0} \mathbf{T}^{c0}_{v} \mathbf{v}_{4\times1}
\end{equation}

where $\mathbf{P}_{c2}$ is projection matrix, $\mathbf{T}^\text{rect}_{c0}$ is transformation matrix from camera $c0$ coordination to rectified camera $c0$ coordination, $\mathbf{T}^{c0}_{v}$ is transformation matrix from lidar $\mathbf{v}$ coordination to camera $c0$ coordination. 
We form the depth projection image $\mathbf{D}_{v}$ from depth projection points $\mathbf{d}_{v}$.

 \begin{figure*}[ht]
    \centering
    \includegraphics[width=14cm]{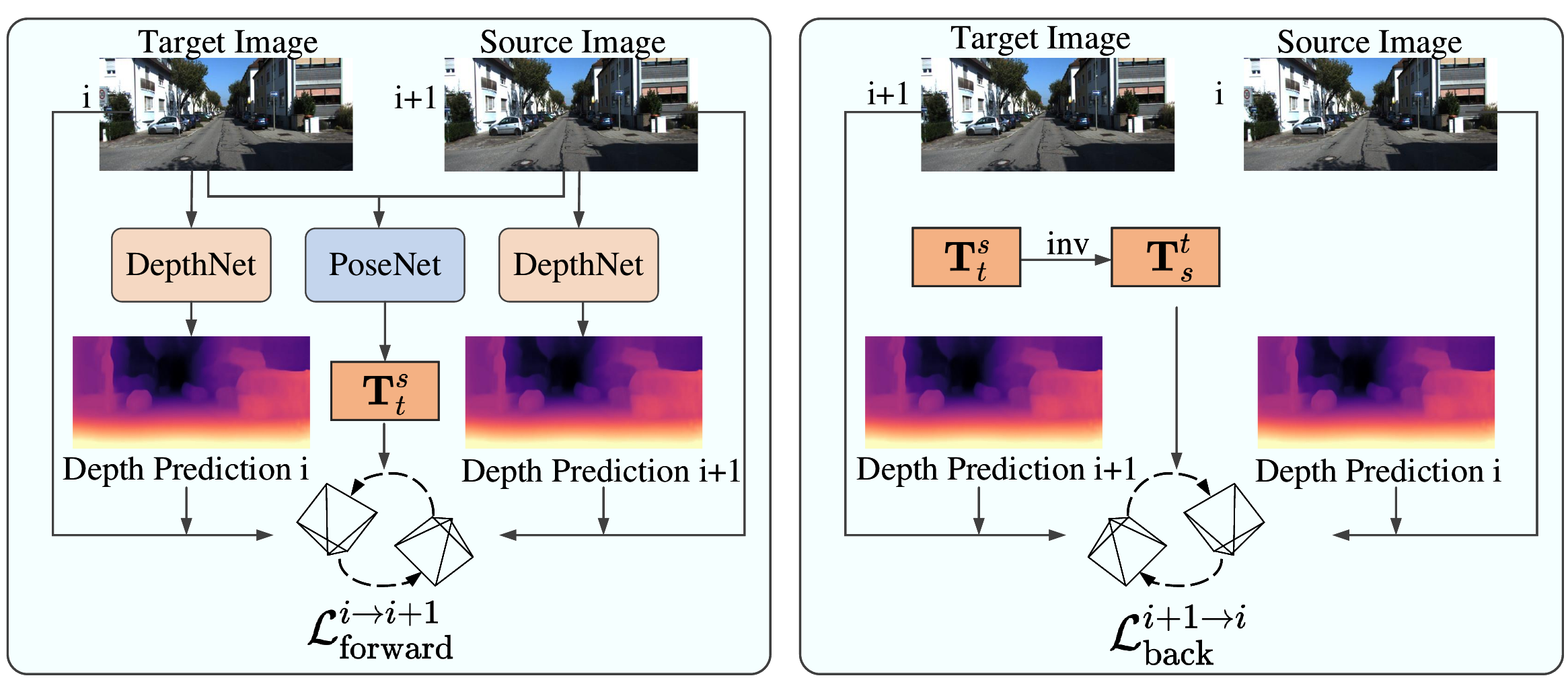}
    \caption{An overview on forward loss and backward loss. }
    \label{fig3:The calculation of bi-directional loss}
\end{figure*}

Then we calculate the ratio between the median value of depth projection images $\text{med}(\mathbf{D}_{v})$ and the median value of depth predictions from pre-trained DepthNet $\text{med}(\mathbf{D}_{t})$. 
The scale factor $\varepsilon$ is calculated as below,
\begin{equation}
    \label{eq: scale factor}
    \varepsilon = \frac{\text{med}(\mathbf{D}_{v})}{\text{med}(\mathbf{D}_{t})}
\end{equation}

Finally, the scale factor $\varepsilon$ is multiplied with the depth predictions to obtain the depths with absolute scale $\bar{\mathbf{D}}$. An example of coarse depth scale recovery is shown in figure \ref{fig2:Generation of depth labels}.
\begin{equation}
    \label{eq: coarse depth}
    \bar{\mathbf{D}} = \varepsilon \cdot  \mathbf{D}
\end{equation}

\subsection{Single-directional Coarse Pose Scale Recovery}
Our self-supervised learning framework leverages depths with absolute scale to recover the pose scale. We propose a two-stage process to achieve this. The first stage estimates coarse poses by utilizing the calculated coarse depths, while the second stage refines the network estimates by jointly optimizing poses and depths.

Firstly, coarse poses are obtained by integrating the calculated coarse depths from Equation \ref{eq: coarse depth} into the Equation \ref{eq: photometric loss} to produce the synthesized images $\hat{\mathbf{I}}_s$ in target frame. We calculate the difference between the wrapped images $\hat{\mathbf{I}}_s$ and the real image in target frame $\mathbf{I}_t$ to construct a photometric loss. Besides, SSIM loss is added to mitigate complex illumination changes \cite{SSIM}. We formulate our photometric loss $L_\text{photo}$ as below,
    \begin{equation}
        \label{deqn_pho}
        \mathcal{L}_\text{pho}=\lambda_1 \cdot |\mathbf{I}_t-\mathbf{\hat{I}}_s| + \lambda_2 \cdot \frac{1-\text{SSIM}(\mathbf{I}_t, \mathbf{\hat{I}}_s)}{2},
    \end{equation}
where $\lambda_1$ and $\lambda_2$ are set as 0.15 and 0.85.
The specific function of SSIM can refer to \cite{SSIM}.

To encourage pose and depth predictions to be scale consistent, a 3D geometric consistency loss  $L_\text{GC}$\cite{bian2019unsupervised}  {between two adjacent images} is introduced: 
    \begin{equation}
        \label{deqn_GC}
        \mathcal{L}_\text{GC} =\frac{|\mathbf{\hat{D}}_s-\mathbf{T}_{t}^{s} \mathbf{D}_t|}{\mathbf{\hat{D}}_s+\mathbf{T}_{t}^{s} \mathbf{D}_t},
    \end{equation}
where $\mathbf{T}_{t}^{s} \mathbf{D}_t$ indicates the process of wrapping the target depth $\mathbf{D}_t$ into the source frame view using the relative pose $\mathbf{T}_{t}^{s}$. And $\mathbf{\hat{D}}_s$ is the interpolated source depth map aligning with $\mathbf{T}_{t}^{s} \mathbf{D}_t$.   

Many unsupervised ego-motion networks in existence optimize the learned pose and depth between adjacent frames. However, our framework takes into account the temporal dependency of poses and depths in consecutive frames and optimizes them in a sequence window. This further encourages the pose scale-recovery to be smoother in a larger window size. 
We set $i$ as the time step of the target frame, $i+1$ as the time step of source frame, and $n$ as the sequence window length. In each time step, we combine photometric loss $\mathcal{L}^{i}_\text{pho}$ and 3D geometric consistency loss $\mathcal{L}^{i}_\text{GC}$ to form an optimization loss. 
Combining the total losses of $n$ frames, we form the single directional scale-recovery loss $\mathcal{L}^{i\to i+1,i\in n-1}$ as below,
\begin{equation}
    \label{deqn_positive}
    \mathcal{L}^{i\to i+1}_\text{forward}
    =\lambda_3\cdot \sum_{i=1}^{n}{\mathcal{L}^{i}_\text{pho}} + \lambda_4\cdot \sum_{i=1}^{n}{\mathcal{L}^{i}_\text{GC}}
\end{equation}
where $\lambda_3$ and $\lambda_4$ are set as 1 and 0.5. 

We use the calculated coarse depths $(\bar{\mathbf{D}}_t,\bar{\mathbf{D}}_s)$ as the target and source depth in this stage to construct a coarse scale recovery loss $\mathcal{L}_\text{coarse}$,
\begin{equation}
\label{deqn_1st.stage}
\mathcal{L}_\text{coarse}=\mathcal{L}^{i\to i+1}_\text{forward}|_{(\bar{\mathbf{D}}_t,\bar{\mathbf{D}}_s)}  
\end{equation}

The model is trained for several epochs to obtain a PoseNet that recovers the coarse scale. Afterward, the entire framework is further refined in the next stage by jointly optimizing PoseNet and DepthNet together.
 
 \subsection{Bi-directional Coarse-to-Fine Pose Scale Recovery}
In the coarse-scale recovery stage, depths are determined through a combination of a pre-trained model and an 
estimated scale factor. This stage enables joint learning of pose and depth, leading to further refinement of estimates for both parameters.
Rather than relying solely on calculated depths, we incorporate depth predictions generated by DepthNet as both the target and source depths when constructing the photometric loss and 3D geometry consistency loss for the coarse-to-fine scale recovery process. 
By optimizing in both forward and backward directions, the model avoids the issue of overfitting on one side, which could lead to unstable training. Therefore, we introduce a backward loss,
\begin{equation}
\label{deqn_negative}
\mathcal{L}^{i+1\to i}_\text{back}
=\lambda_3\cdot \sum_{i=1}^{n}{\mathcal{L}^{i}_\text{pho}} + \lambda_4\cdot \sum_{i=1}^{n}{\mathcal{L}^{i}_\text{GC}}
\end{equation}
Here, as shown in Figure \ref{fig3:The calculation of bi-directional loss}, the relative pose and the order of target and source frame are inverse.

 {During the coarse-scale recovery stage, only the parameters of the PoseNet need to be optimized, while the pseudo depth labels remain fixed and non-learnable. To provide robust training, we employ single-directional photometric and geometric losses as strong training constraints. However, during the coarse-to-fine scale recovery stage, both the PoseNet and DepthNet parameters need to be optimized without the assistance of pseudo depth supervision, and the DepthNet is trained from scratch. This leads to learnable depth estimates that introduce uncertainty to the training process. To mitigate this training uncertainty, we introduce additional photometric and geometric losses in the backward direction as training constraints. In our experiments, we observe that applying only a single-directional training loss during the second stage results in overfitting to one direction, leading to unstable training and meaningless estimates. Therefore, we introduce more training constraints to promote convergence of the self-supervised framework training and facilitate optimal performance. Compared to fixed and non-learnable pseudo depth labels, learnable depth estimates offer greater potential for loss reduction, which promotes network training convergence to optimal performance. In summary, bi-directional scale recovery loss in the second stage is necessary to ensure that the framework works as intended and further improves its performance.}

Furthermore, we introduce an edge-aware smoothness loss $\mathcal{L}_\text{smooth}$ to overcome the shortage of photometric loss in the low-texture regions \cite{Smooth}.
\begin{equation}
\label{deqn_ex2a}
 {\mathcal{L}_\text{smooth} = \sum_{p\in(u,v)}^{} (e^{-\bigtriangledown \mathbf{I}_t(p)}\cdot \bigtriangledown\mathbf{D}_t(p))^2}
\end{equation}
where $p$ indicates pixel point in coordination $(u,v)$, and $\mathbf{I}_t(p)$ indicates the pixel value in target frame $\mathbf{I}_t$, $\bigtriangledown$ means the first derivative along two spatial directions on pixel coordination $(u,v)$.

Combining the forward $\mathcal{L}^{i\to i+1}$ loss, backward $\mathcal{L}^{i+1 \to i}$ loss and smoothness loss $\mathcal{L}_\text{smooth}$, the total loss for bi-directional scale recovery is formulated as:
\begin{equation}
\label{deqn_ex2a2}
\mathcal{L}_\text{refine}=\lambda_5(\mathcal{L}^{i\to i+1}_\text{forward}  +\mathcal{L}^{i+1\to i}_\text{back})|_{({\mathbf{D}_t},{\mathbf{D}_s})} +\lambda_6\cdot\mathcal{L}_\text{smooth}
\end{equation}

where we set $\lambda_5=1$ and $\lambda_6=0.1$. ${({\mathbf{D}_t},{\mathbf{D}_s})}$ are the learned depth from DepthNet.  {The $\lambda_1\sim\lambda_6$ mentioned above are all non-learnable hyperparameters}.

\subsection{Discussion}
We also attempted to train the overall framework directly using coarse-scaled depth predictions, rather than dividing the training process into two stages: coarse scale recovery and coarse-to-fine scale recovery. However, our experimental results indicated that introducing depth labels alone was insufficient for recovering pose and depth scale directly. We attribute this limitation to the non-learnable nature of absolute-scale depth supervisions. Specifically, at the outset of training, the photometric loss associated with depth supervision was large, while the loss reduction potential was small. In contrast, the photometric loss formed by the learnable depth prediction decreased with each training iteration, with a larger loss reduction potential. Additionally, the direction of network training was consistent with the direction of loss reduction. As a result, scaled depth supervision did not facilitate the framework's ability to recover absolute scale through direct training.  {Therefore, we employ a two-stage training process. In the first stage, we use the scaled pseudo depth label to provide training supervision. This label introduces the absolute scale factor to the photometric and geometric loss functions, enabling PoseNet to recover the scale of pose estimates during the training process. In the second stage, we finetune PoseNet with a smaller learning rate to prevent it from forgetting the scale information, even without the pseudo depth labels. Since the pose estimates of PoseNet have already been scale recovered, the scale factor learned in the first stage is introduced into the second stage, enabling the DepthNet to recover the scale of pose estimates during the second training process. The scaled pseudo depth label, which comes from the well-pretrained DepthNet, allows PoseNet to converge quickly and recover the scale of pose estimates at around 45 epochs in the first stage. In the second stage, we initialize PoseNet with the parameters trained in the first stage, enabling it to converge rapidly. It takes about 50 epochs to achieve optimal pose estimation performance. Since we train the DepthNet from scratch in the second stage, it takes about 90 epochs to reach optimal depth estimation performance.}

\section{Experiment}
This section introduces implementation details, and discusses the evaluation of pose and depth estimation of our proposed model above two datasets. In addition, ablation study is conducted to verify the effectiveness of the important modules in our model.

\begin{figure}[htb]
\centering
\subfloat[Day-time  image]{\includegraphics[width=0.48\linewidth]{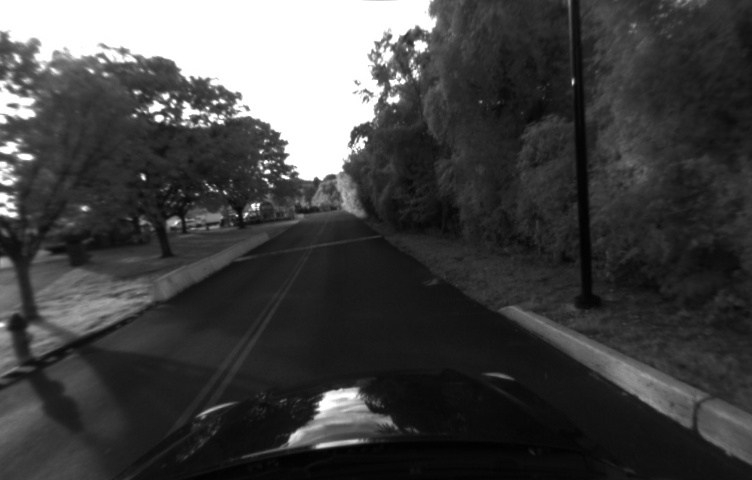}}
\hfill
\subfloat[Night-time image]{\includegraphics[width=0.48\linewidth]{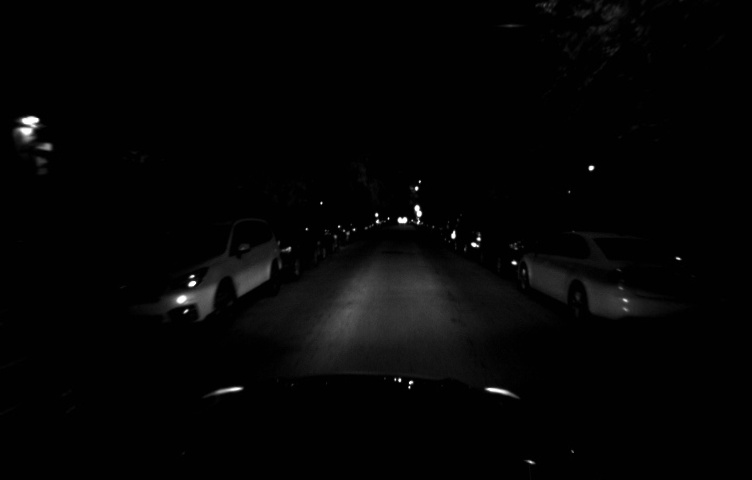}}
\caption{The sample images of the MVSEC dataset: a) driving in daytime b) driving at night}  
\label{MVSEC}
\end{figure}

\subsection{Training details} 
We use PyTorch to implement our proposed network. Our model is trained and tested via a NVIDIA RTX3090 GPU. Adam is chosen as optimizer to recover the optimal parameters, whose attenuation coefficient is $\beta _1 = 0.9$,  $\beta _2 = 0.999$. The model is trained for 200 epochs, and in each epoch there are 1000 random data sequences. 
The model training follows a two-stages process: first, we use the pseudo depth labels to train the network for around 30-100 epochs, until a coarse pose estimation with absolute scale can be recovered; then, depth and pose networks are jointly trained to enable coarse-to-fine pose and depth estimation. The learning rate is chose as 1e-4 in the first stage and reduced to 1e-5 in the second stage.

\begin{figure}[!t]
	\centering
	\subfloat[KITTI Sequence 09]{\includegraphics[width=9.2cm]{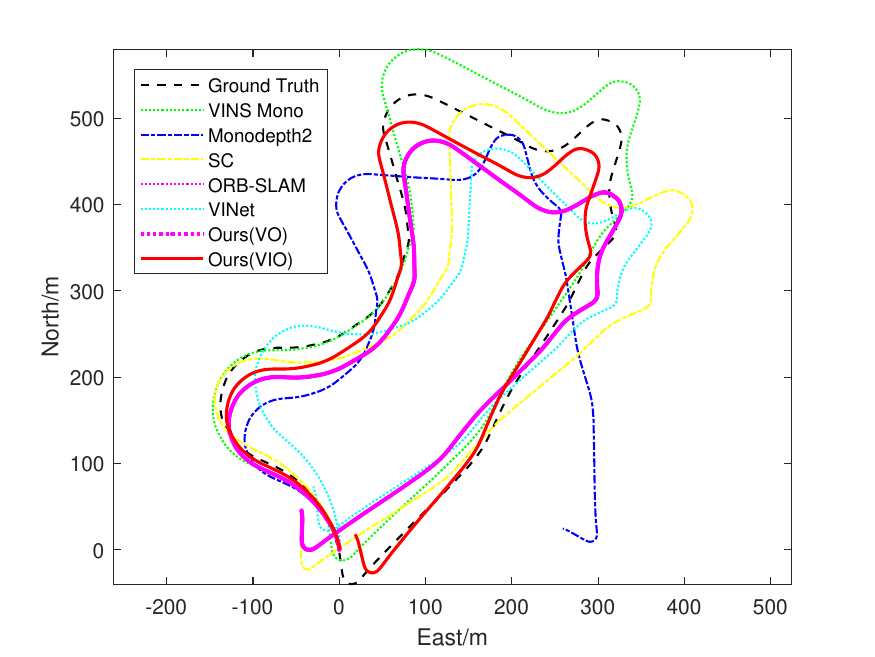}%
		\label{fig_first_case}}
	\hfil
	\subfloat[KITTI Sequence 10]{\includegraphics[width=9.2cm]{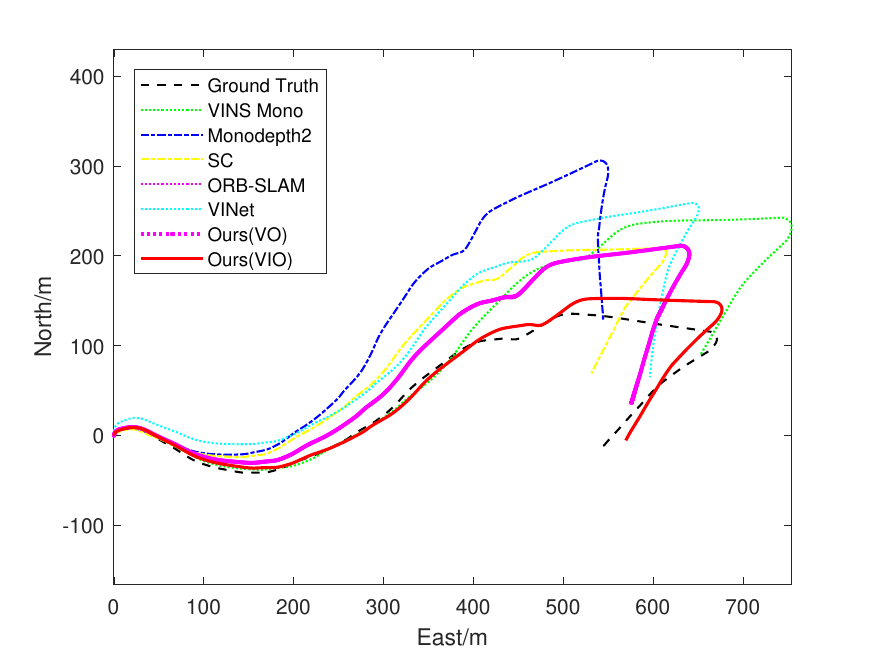}%
		\label{fig_second_case}}
	\caption{The trajectories of our proposed self-supervised VO and VIO model on Sequence 09 and 10 of the KITTI RAW dataset, comparing with baselines. }
	\label{fig: trajectory kitti}
\end{figure}

\subsection{Datasets}
\subsubsection{KITTI Odometry Dataset}
The KITTI Raw dataset is a widely used dataset for car-driving scenarios \cite{KITTI}, providing RGB images, IMU, LIDAR point cloud data, and ground-truth pose data. We use this dataset to generate pseudo depth labels and training data for our odometry network. The dataset includes 10 sequences, with Sequence 03 excluded due to missing IMU data. We use Sequences 00-08 for training and evaluation, and Sequences 09 and 10 for testing. To synchronize the unsynchronized IMU and LIDAR data, we manually align these data according to their timestamps. The IMU data has a frequency of 100Hz, while the images and synchronized LIDAR point cloud data are captured at a frequency of 10Hz. The ground-truth pose is used only for testing, to evaluate the pose accuracy, and is not involved in the training process.

\subsubsection{MVSEC Dataset}
To evaluate the performance of our model in low-light conditions, we utilize the Multi Vehicle Stereo Event Camera (MVSEC) dataset \cite{MVSEC}, which includes gray images and IMU data. The images are sampled at a frequency of 20Hz, while IMU data are recorded at a frequency of 200Hz. Our experiments employ data from car-driving scenarios in both daytime and nighttime, consisting of two daytime sequences (Day1, Day2) and three nighttime sequences (Evening1, Evening2, Evening3). We use Day1, Day2, Evening1, and Evening2 as training data, and Evening3 as testing data. As illustrated in Figure \ref{MVSEC}, the low-light conditions at night pose challenges to visual pose and depth estimation. The dataset also provides ground-truth pose, which we use solely for testing and evaluation.

\subsection{Pose evaluation on the KITTI dataset} 
In order to evaluate the effectiveness of our proposed VO/VIO models, we conducted experiments on the public KITTI dataset. To assess the performance of our models, we used the official KITTI evaluation, which computes the root-mean-square error (RMSE) of the translation and rotation vectors over sequences ranging from 100m to 800m in length, and averages these values to obtain an overall pose accuracy metric.

\begin{table*}
\centering
\small
\caption{The pose performance on the KITTI dataset. "GT" indicates  whether the translation estimates are scaled by ground truth. M and I indicate monocular camera and IMU respectively.}
\renewcommand\arraystretch{1.3} 
\begin{tabular}{ccccccccc}
\hline
\multicolumn{1}{c}{\multirow{2}{*}{Method}} & \multirow{2}{*}{Sensors} & \multirow{2}{*}{GT}  & \multicolumn{2}{c}{Seq. 09} & \multicolumn{2}{c}{Seq. 10} & \multicolumn{2}{c}{Avg} \\ \cline{4-9} & & & $t_{rel}$   & $r_{rel}$   & $t_{rel}$   & $r_{rel}$   & $t_{rel}$ & $r_{rel}$ \\ \hline
ORB-SLAM      & Mono  &  $\checkmark$     & 15.3         & \textbf{0.26}         & 3.68         & \textbf{0.48}         & 9.49       & \textbf{0.37}       \\ 
Depth-VO-Feat & Stereo & x   & 11.92& 3.60 &12.62 &3.43 &12.27 &3.515 \\ 
GeoNet & Mono & $\checkmark$   &23.94 &9.81 &20.73 &9.10 &22.34 &9.46 \\
Monodepth2  &Mono    & $\checkmark$          & 18.12        & 3.86         & 12           & 5.34         & 15.06      & 4.6        \\ 
SfMLearner &Mono & $\checkmark$   &17.84 &6.78 &37.91 &17.78 &27.875 &12.28 \\ 
SC           &Mono & $\checkmark$          & 8.62         & 3.05         & 7.81         & 4.9          & 8.22       & 3.98       \\ 
Wagstaff et al.  &Mono & x           & \textbf{5.93}         & 1.67         & 10.54         & 4.03          & 8.23       & 2.85       \\ 
SelfOdom(VO)     &Mono & x               &  7.90         &  1.23        &  \textbf{6.55}       &  2.05        &  \textbf{7.23}      &  1.64       \\
\hline
VINS        &M+I  & x       & 41.47        & 2.41         & 20.35        & 2.73         & 30.91      & 2.57       \\ 
VINet          &M+I   & x      & 11.83        & 3.00         & 8.60         & 4.39         & 10.22      & 3.70       \\ 
UnVIO &M+I & $\checkmark$  & 4.41 & 0.92 & 6.42 & 0.63 & 5.42 & 0.78 \\ 
SelfOdom(VIO) &M+I & x  & 5.48 & 0.19 & 5.37 & 0.43  & 5.43  & 0.31 \\
SelfOdom(VIO) & M+I & $\checkmark$    & \textbf{3.65} & \textbf{0.19} & \textbf{4.42} & \textbf{0.43}  & \textbf{4.04}  & \textbf{0.31} \\\hline
\end{tabular}
\label{tb: pose kitti}
\end{table*}

\begin{table*}
\centering
\caption{The depth evaluation on the KITTI dataset. These models are trained with Sequence 00-08, and tested with Sequence 09 and 10. "Scale" indicates whether the depth estimates are scaled or not.}
\renewcommand\arraystretch{1.5}
\begin{tabular}{cccccccccc}
\hline
\multicolumn{1}{c}{\multirow{2}{*}{Method}} &\multirow{2}{*}{Sensors} & \multirow{2}{*}{Resolution}&\multirow{2}{*}{Scale} &\multicolumn{3}{c}{Error metric $\downarrow$ } & \multicolumn{3}{c}{Accuracy metric($\delta$ $\uparrow$)}                                               \\ \cline{5-10} 
             &    &  &     & Abs Rel   & Sq Rel   & RMSE  & (1.25)   & (1.25\textasciicircum{}2)  & (1.25\textasciicircum{}3) \\ \hline
SfMLearner   &Mono    &$832\times256$  & x     & 0.3272    & 3.1131    & 9.5216   & 0.4232          & 0.7010                             & 0.8476                             \\ 
SC           &Mono   &$832\times256$   & x     & 0.1629    & 0.9644    & 4.9129   & 0.7760          & 0.9315                             & 0.9773                             \\ 
UnVIO        &Mono+IMU   &$832\times256$   & x     & 0.1322    & 0.73005   & \textbf{4.2443}   & 0.8324          & 0.9509                             & 0.9821                             \\ 
Ours1 &Mono &$832\times256$   & x & 0.1322   & 0.7244    & 4.2715   & 0.8313          & 0.9498                             & 0.9816                             \\ 
Ours2 &Mono  &$832\times256$   & \checkmark  & 0.1308    & 0.6980    & 4.2930   & 0.8318          & 0.9520                             & 0.9836                             \\
Ours3 &Mono+IMU  &$832\times256$   & \checkmark   & \textbf{0.1301}    & \textbf{0.6942}    & 4.2963   & \textbf{0.8341}  & \textbf{0.9526}  & \textbf{0.9837} \\ 
\hline
\end{tabular}
\label{tb: depth kitti}
\end{table*}

For the vision-only pose estimation task, we compared our model against a traditional SLAM method (ORB-SLAM \cite{ORB}), as well as five learning-based VO models, including Depth-VO-Feat \cite{Depth-vo}, GeoNet \cite{GeoNet}, MonoDepth2, SfMLearner \cite{SfMLearner}, SC \cite{bian2019unsupervised} and Wagstaff et al.\cite{SelfSupervisedSR}. Among these, GeoNet, MonoDepth2, and SfMLearner are scale-ambiguous, while SC is scale-consistent, but still lacks absolute scale. Wagstaff et al.\cite{SelfSupervisedSR} recovers the absolute scale by employing a predefined camera height. In contrast, our proposed self-supervised VO model is capable of producing pose estimates with global scale without the need of predefined parameters. The results, presented in Table \ref{tb: pose kitti}, demonstrate that our VO model outperforms all traditional and learning-based VO baselines in terms of translation, and outperforms all learning-based VO models in terms of rotation. Moreover, our VO model improves the rotation estimation of SC by 58.79\%, on average, and improves the translation estimation by 12.04\%. 
Pure visual pose estimation methods often struggle to extract sufficient visual features in low lighting and texture-less environments, which can result in decreased pose estimation performance. To overcome these challenges, SelfOdom integrates the motion measurement information from the gyroscope and accelerometer, which is independent of external visual information. In this approach, we utilize the inertial encoder to extract inertial features from the IMU data and fuse these with the visual features using a soft attention module. By incorporating inertial data into our learning-based VIO model, we demonstrate superior performance in rotation estimation compared to all VO/VIO baselines. This combination of visual and inertial data enables our model to overcome the adverse effects of complex lighting environments, resulting in more accurate and reliable pose estimates.

In order to assess the efficacy of our proposed VIO model, we conducted a comparative analysis against three other state-of-the-art VIO methods, including a traditional VIO method, VINS, and two learning-based VIO models, VINet and UnVIO. VINet is a supervised learning-based VIO model that employs a combination of ConvNet and LSTM, while UnVIO is an unsupervised learning-based VIO model that leverages novel view synthesis techniques similar to our model, but lacks an absolute scale metric. The quantitative results presented in Table \ref{tb: pose kitti} demonstrate that our VIO model significantly outperforms these baselines in terms of both translation and rotation estimation. Specifically, our VIO model improves the translation and rotation estimation of UnVIO by 25.46\% and 60.26\%, respectively. Furthermore, our VIO model is equipped with an absolute scale metric, which enables it to produce comparable performance even when ground truth is not available to scale translation estimates. Visually, the generated trajectories from our VO and VIO models, as depicted in Figure \ref{fig: trajectory kitti}, closely approximate the ground-truth trajectories, outperforming other baselines, including VINS-mono, MonoDepth2, SC, ORB-SLAM, and VINet.

\begin{figure*}
\centering
\begin{minipage}[b]{0.08\linewidth} 
\centering 
\captionsetup{font={small}}
\caption*{Input}\vspace{10pt}
\caption*{SfM}\vspace{12pt}
\caption*{SC}\vspace{9pt}
\caption*{Ours(VO)}\vspace{9pt}
\caption*{Ours(VIO)}
\end{minipage}
\begin{minipage}[b]{0.15\linewidth}
\includegraphics[width=1\linewidth]{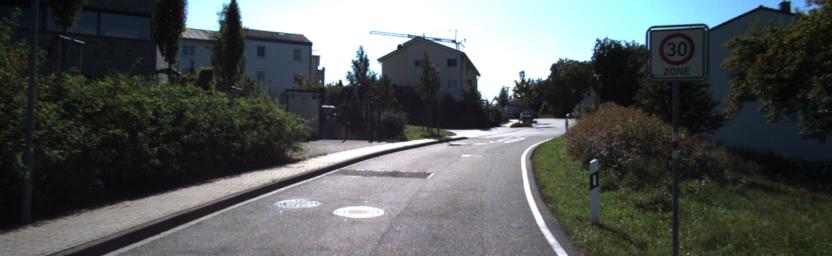}\vspace{4pt}
\includegraphics[width=1\linewidth]{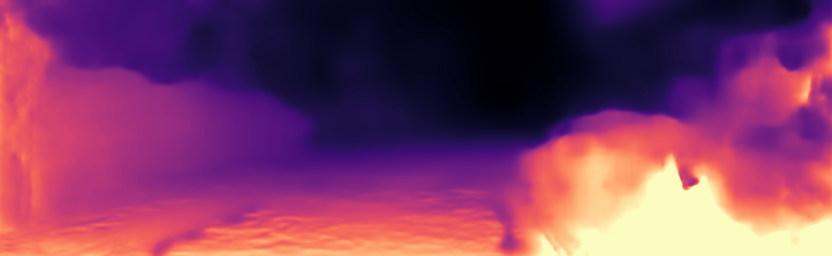}\vspace{4pt}
\includegraphics[width=1\linewidth]{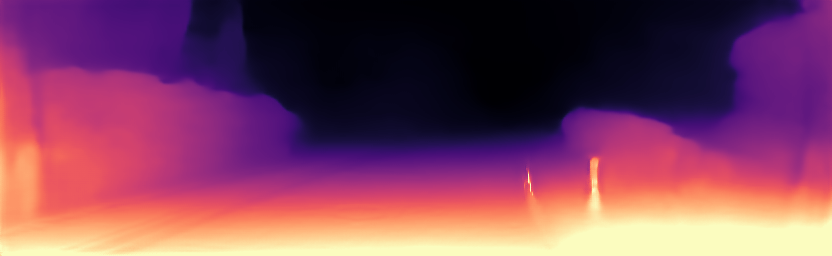}\vspace{4pt}
\includegraphics[width=1\linewidth]{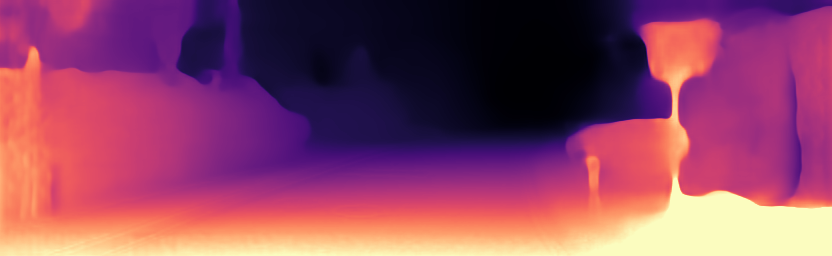}
\includegraphics[width=1\linewidth]{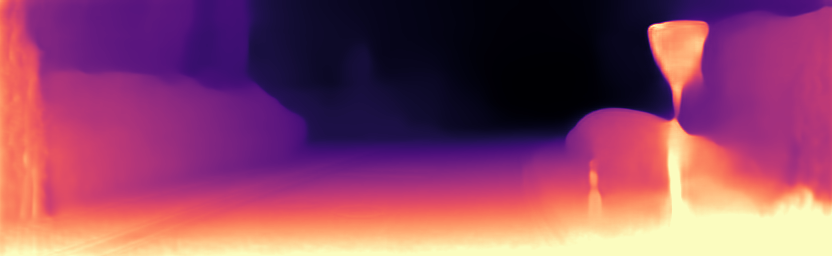}
\end{minipage}
\begin{minipage}[b]{0.15\linewidth}
\includegraphics[width=1\linewidth]{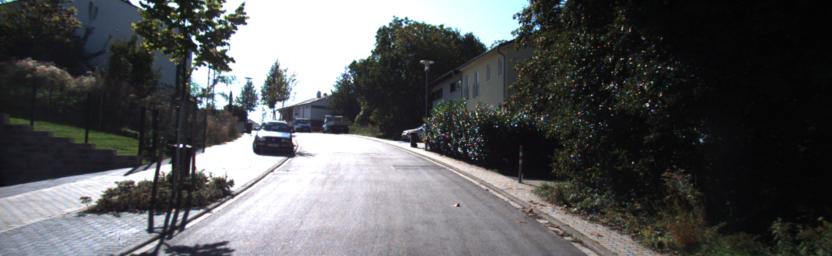}\vspace{4pt}
\includegraphics[width=1\linewidth]{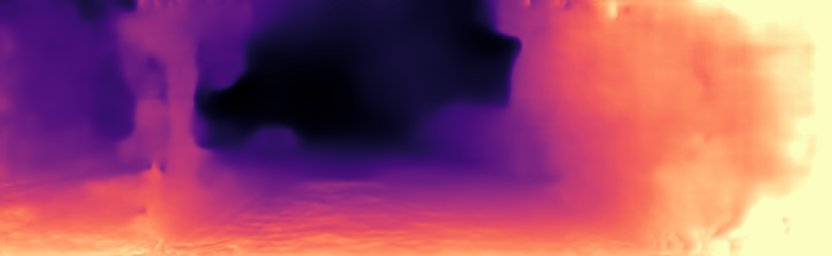}\vspace{4pt}
\includegraphics[width=1\linewidth]{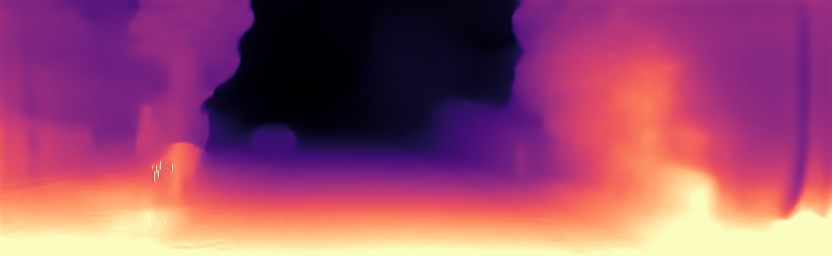}\vspace{4pt}
\includegraphics[width=1\linewidth]{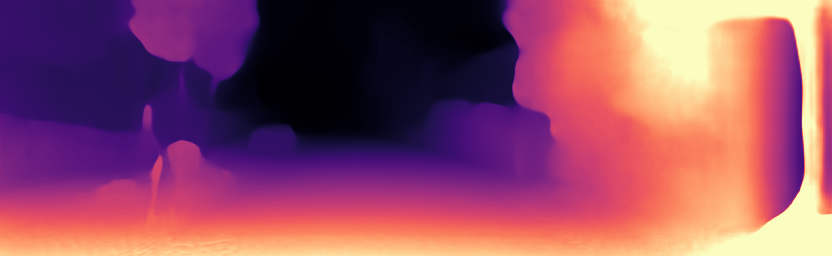}
\includegraphics[width=1\linewidth]{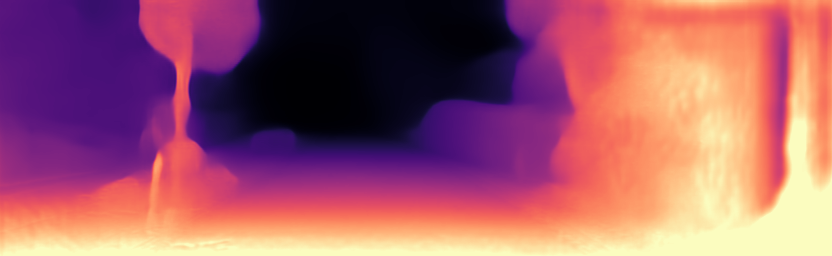}
\end{minipage}
\begin{minipage}[b]{0.15\linewidth}
\includegraphics[width=1\linewidth]{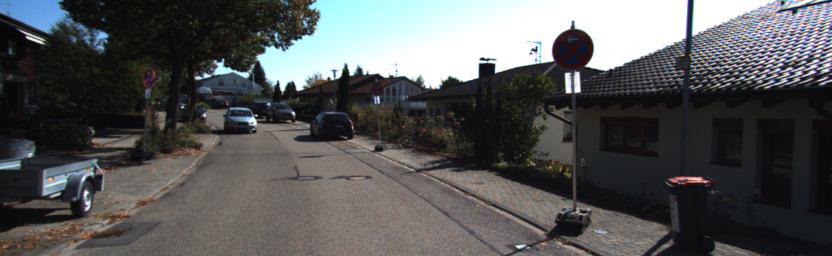}\vspace{4pt}
\includegraphics[width=1\linewidth]{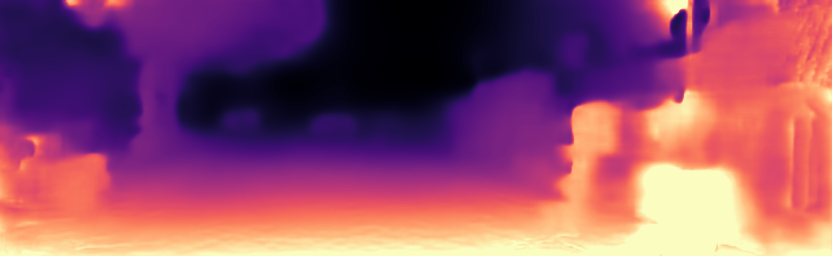}\vspace{4pt}
\includegraphics[width=1\linewidth]{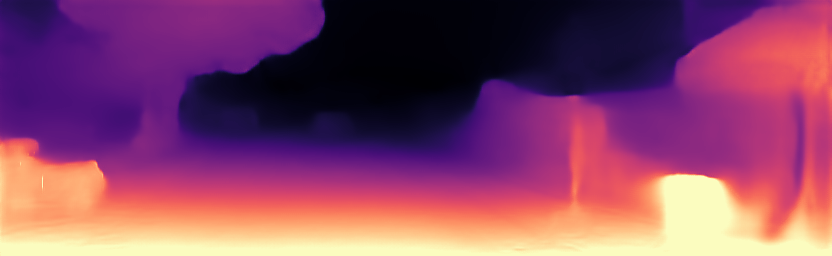}\vspace{4pt}
\includegraphics[width=1\linewidth]{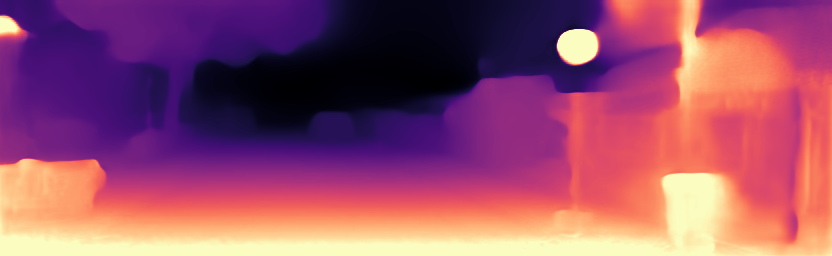}
\includegraphics[width=1\linewidth]{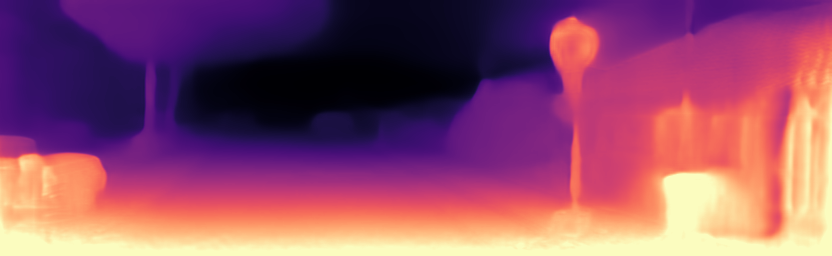}
\end{minipage}
\begin{minipage}[b]{0.15\linewidth}
\includegraphics[width=1\linewidth]{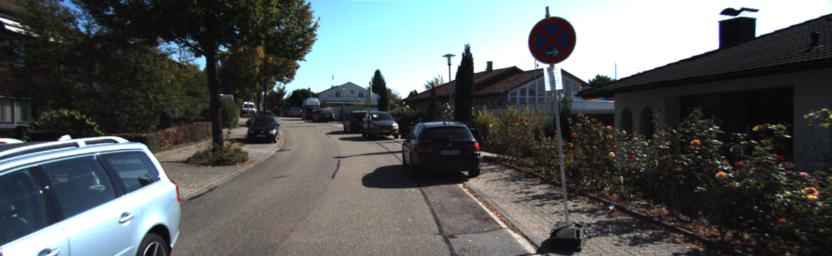}\vspace{4pt}
\includegraphics[width=1\linewidth]{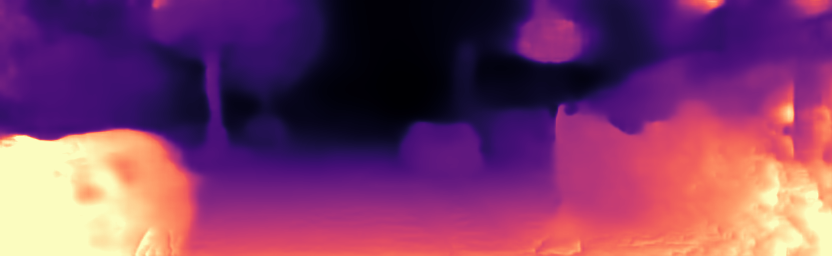}\vspace{4pt}
\includegraphics[width=1\linewidth]{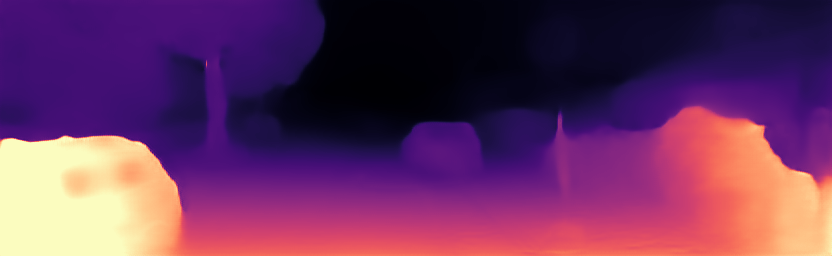}\vspace{4pt}
\includegraphics[width=1\linewidth]{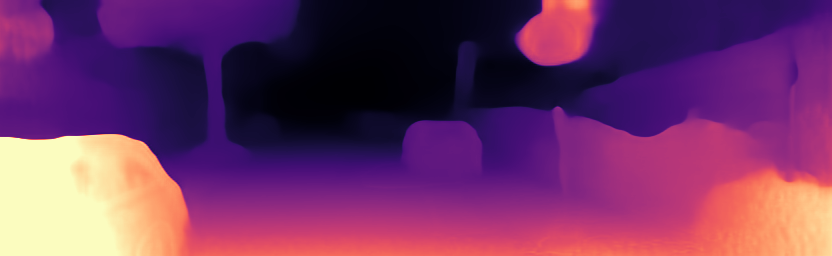}
\includegraphics[width=1\linewidth]{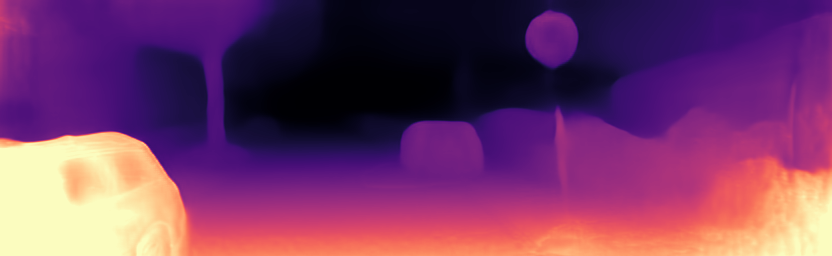}
\end{minipage}
\begin{minipage}[b]{0.15\linewidth}
\includegraphics[width=1\linewidth]{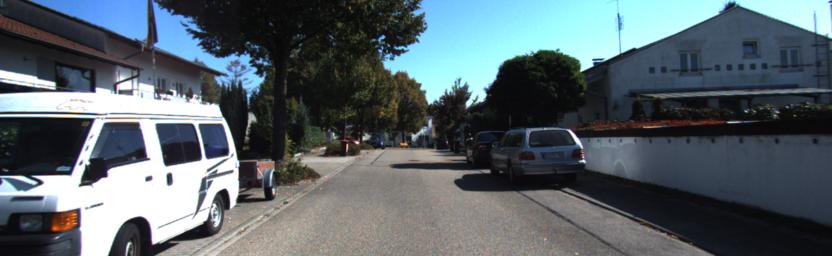}\vspace{4pt}
\includegraphics[width=1\linewidth]{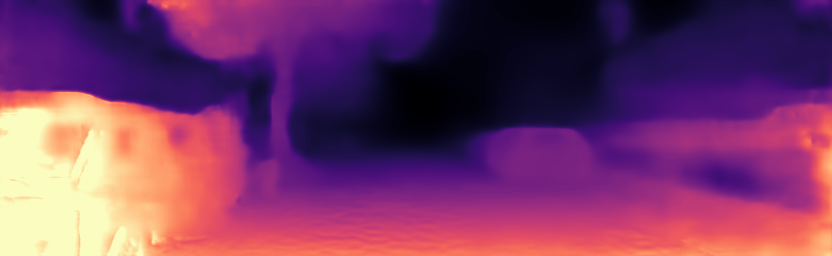}\vspace{4pt}
\includegraphics[width=1\linewidth]{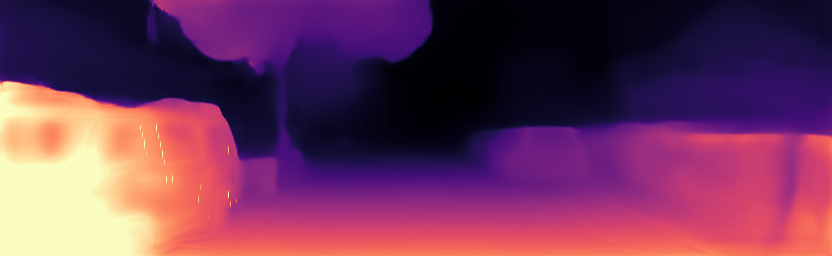}\vspace{4pt}
\includegraphics[width=1\linewidth]{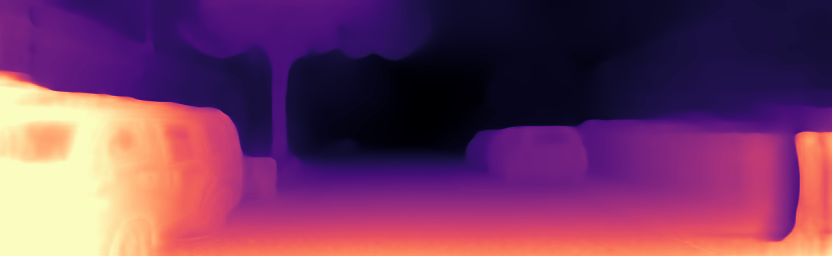}
\includegraphics[width=1\linewidth]{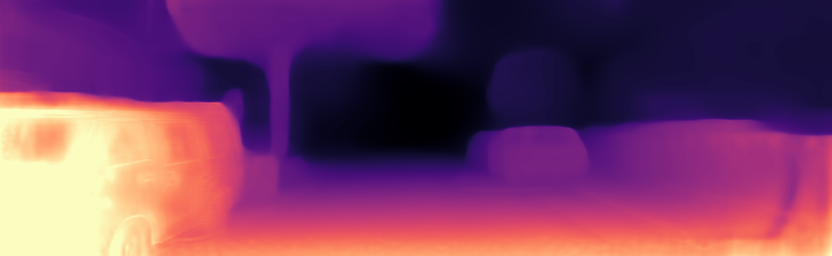}
\end{minipage}
\caption{The samples of depth estimation from SfM-Learner, SC and our proposed models}
\label{fig: depth kitti images}
\end{figure*}

\subsection{Depth evaluation on the KITTI Dataset} 
In addition to evaluating pose estimation, we also conducted experiments to evaluate the depth estimation performance of our proposed models on the KITTI dataset. Unlike SC-SfMLearner, which uses the KITTI raw dataset for training, we trained our models on Sequence 00-08 of the KITTI dataset as these sequences include IMU data. The quantitative results of the depth estimation for our proposed models on Sequence 09 and 10, compared to three mainstream learning-based depth estimation methods, SfMLearner, SC, and UnVIO, are presented in Table \ref{tb: depth kitti}.

To evaluate the accuracy of the depth estimation, we used the following evaluation metrics, where $p$ indicates a pixel point in coordinate $(u, v)$, $\mathbf{D}$ and $\mathbf{D}_v$ indicate the learned depth prediction from DepthNet and the corresponding depth projection image, respectively. $a$ is the threshold factor, and $N$ is the sum of pixel points in the depth image coordinate.

\begin{equation}
  \label{deqn_depth_metrics}
  \begin{aligned}
\text{Abs}\ \text{Rel} &=\frac{1}{N}\sum _{p}\frac{\left | \mathbf{D} (p)-\mathbf{D}_v (p) \right | }{\mathbf{D}_v (p)} \\
\text{Sq}\ \text{Rel} &=\frac{1}{N}\sum _{p}\frac{ (\mathbf{D} (p)-\mathbf{D}_v (p))^2 }{\mathbf{D}_v (p)} \\
\text{RMSE} &=\sqrt{\frac{1}{N}\sum_p(\mathbf{D} (p)-\mathbf{D}_v (p))  } \\
\delta (a) &= \frac{1}{N}\sum_p{\text{max}(\frac{\mathbf{D} _v(p)}{\mathbf{D} (p)},\frac{\mathbf{D} _(p)}{\mathbf{D}_v (p)} )<a}
  \end{aligned}
\end{equation}

The results suggest that our depth estimation models, namely Ours2 and Ours3, are capable of producing depth with an absolute scale, and they outperform the other three baseline models. The incorporation of IMU data into depth estimation has little effect on the overall performance. Figure \ref{fig: depth kitti images} showcases several depth images generated by our proposed models and the baseline models using monocular camera. As observed in the figure, our models produce depth maps with finer details compared to the other learning-based depth estimation models. For instance, the poles are distinctly visible in our depth maps.

In addition, we investigated the effectiveness of our proposed scale recovery method. The results are presented in Table \ref{tb: scale factor}, where we report the mean value $\mu$ and standard deviation value $\delta$ of the scale factor for depth estimation with and without scale recovery. The table indicates that the average scale coefficient of depth estimation values is closer to 1 when our coarse-to-fine scale recovery method is applied. Without scale recovery, there is a significant discrepancy between the estimated scale-ambiguous depths and real depths. This finding suggests that incorporating scale recovery not only enables PoseNet to generate pose predictions with a global absolute scale but also enhances the ability of DepthNet to produce depth estimates that are closer to the real absolute scale metric.

\begin{table}[h!]
\small
\centering
\caption{The pose estimation results in the night car-driving scenes.}
\setlength{\tabcolsep}{1.8mm}{
\renewcommand{\arraystretch}{1.5}
\begin{tabular}{cccc}
\hline
\multicolumn{1}{c}{Method}  &\multicolumn{1}{c}{Sensors}&\multicolumn{1}{c}{GT} & \multicolumn{1}{c}{Error} \\ \hline
ORB-SLAM3 Mono      & Mono & \checkmark & failed                     \\ 
SC-SfMLearner       & Mono & \checkmark        & 72.05                          \\
DeepVO              & Mono & x    & 39.37                            \\ 
SelfOdom (VO, ours)            & Mono & x        & 35.12 \\ 
SelfOdom (VIO, ours)           & Mono+IMU & x          & \textbf{19.62} \\ 
\hline
\end{tabular}}
\label{tb: pose night}
\end{table}

\begin{figure*}
\centering
\begin{minipage}[b]{0.05\linewidth} 
\centering 
\caption*{Input}\vspace{58pt}
\caption*{GT}\vspace{44pt}
\caption*{SC}\vspace{34pt}
\caption*{Ours}
\end{minipage}
\begin{minipage}[b]{0.18\linewidth}
\includegraphics[width=1\linewidth]{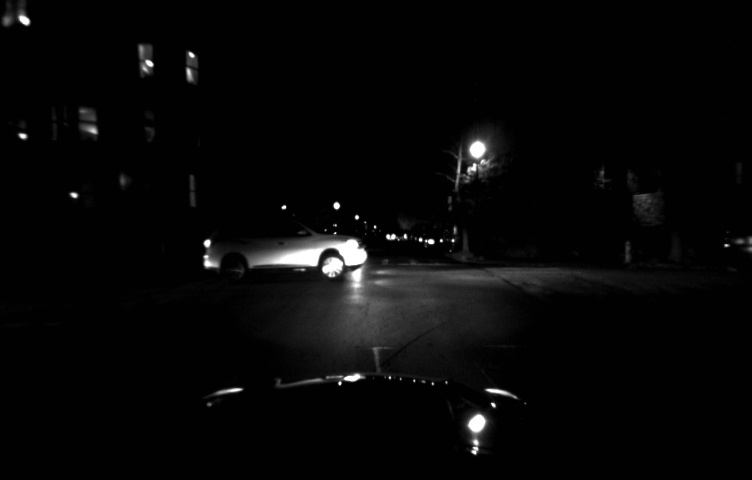}\vspace{4pt}
\includegraphics[width=1\linewidth]{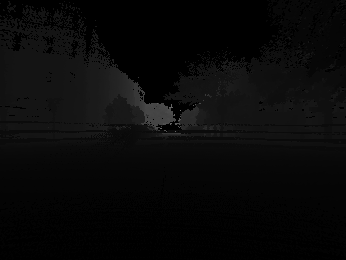}\vspace{4pt}
\includegraphics[width=1\linewidth]{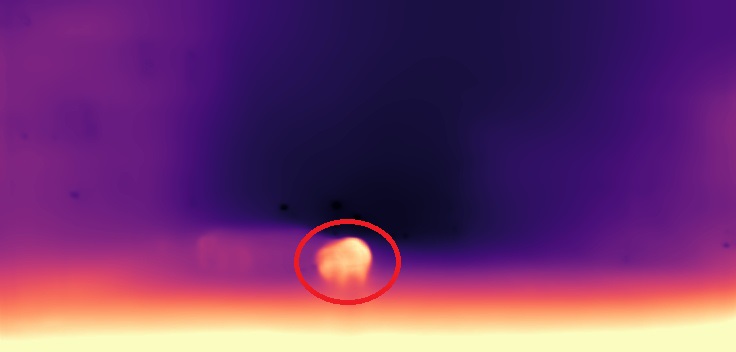}\vspace{4pt}
\includegraphics[width=1\linewidth]{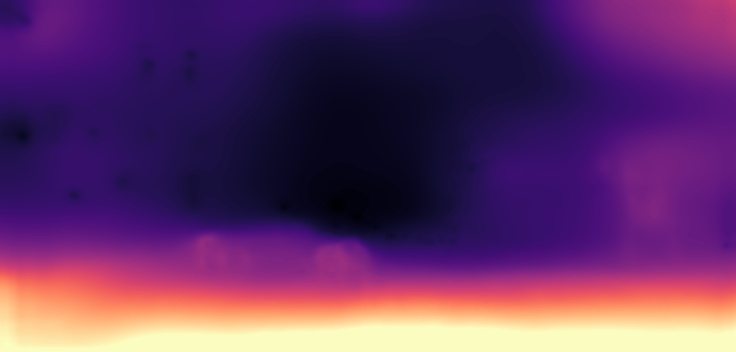}
\end{minipage}
\begin{minipage}[b]{0.18\linewidth}
\includegraphics[width=1\linewidth]{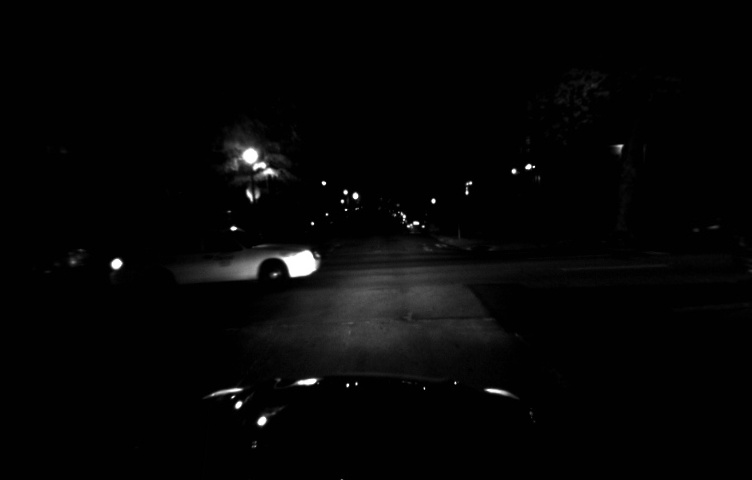}\vspace{4pt}
\includegraphics[width=1\linewidth]{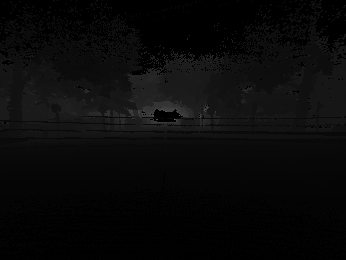}\vspace{4pt}
\includegraphics[width=1\linewidth]{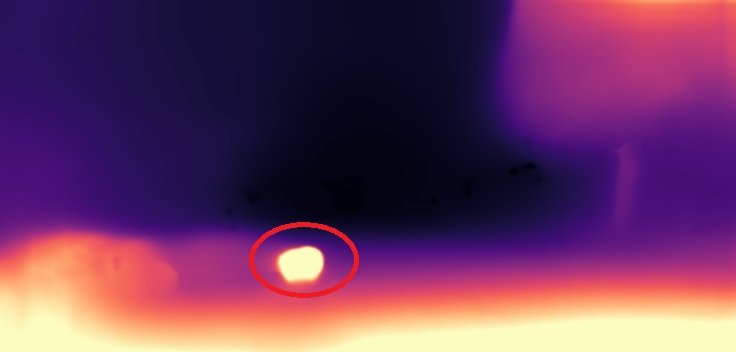}\vspace{4pt}
\includegraphics[width=1\linewidth]{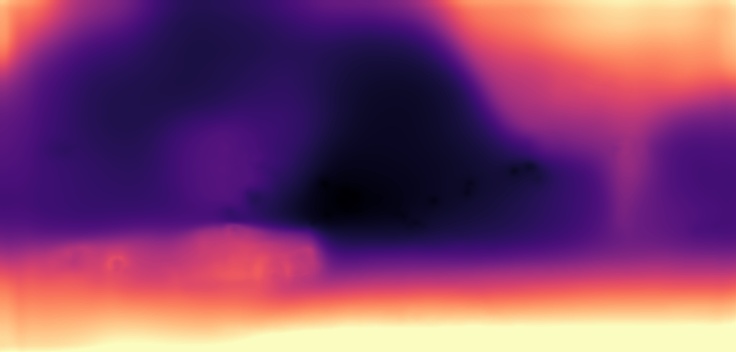}
\end{minipage}
\begin{minipage}[b]{0.18\linewidth}
\includegraphics[width=1\linewidth]{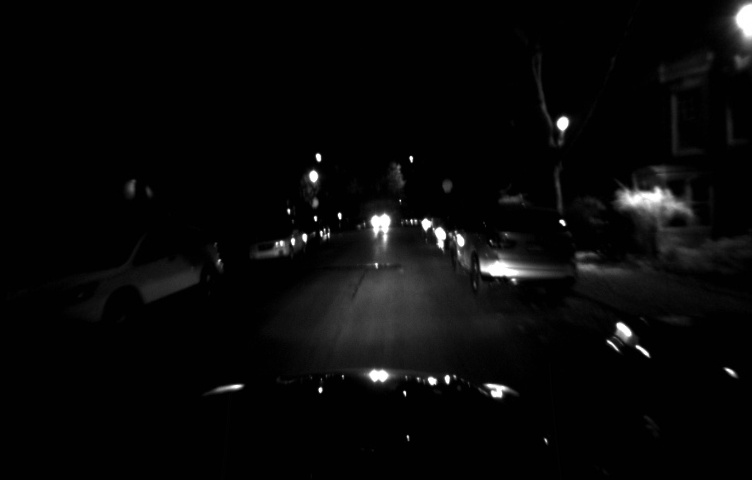}\vspace{4pt}
\includegraphics[width=1\linewidth]{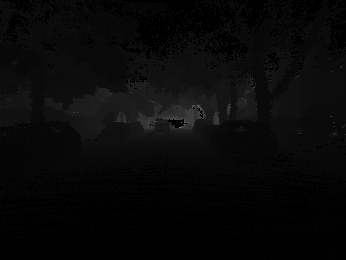}\vspace{4pt}
\includegraphics[width=1\linewidth]{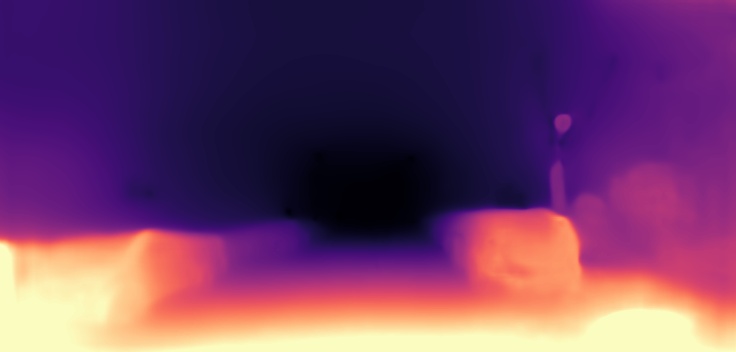}\vspace{4pt}
\includegraphics[width=1\linewidth]{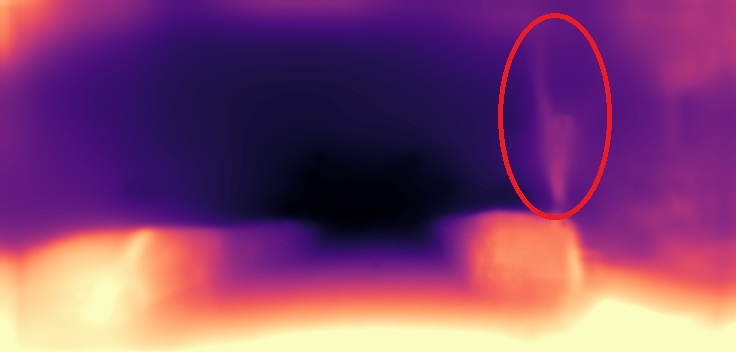}
\end{minipage}
\begin{minipage}[b]{0.18\linewidth}
\includegraphics[width=1\linewidth]{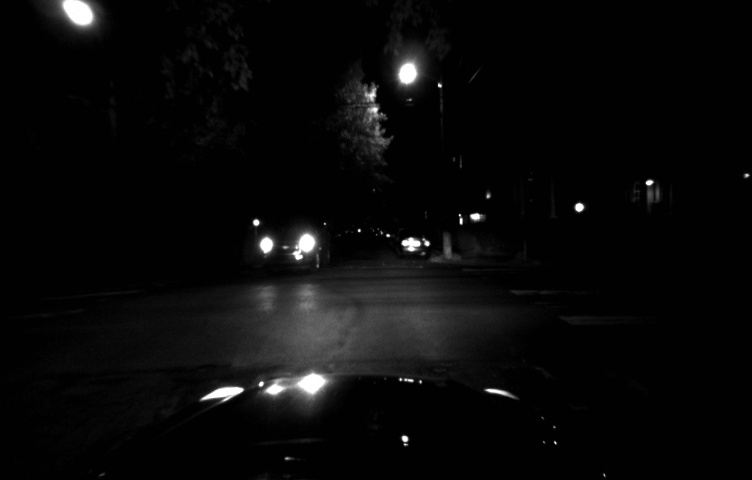}\vspace{4pt}
\includegraphics[width=1\linewidth]{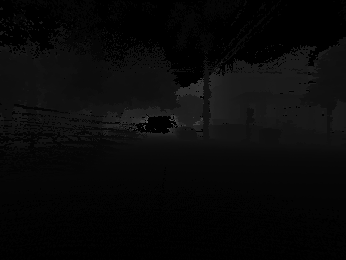}\vspace{4pt}
\includegraphics[width=1\linewidth]{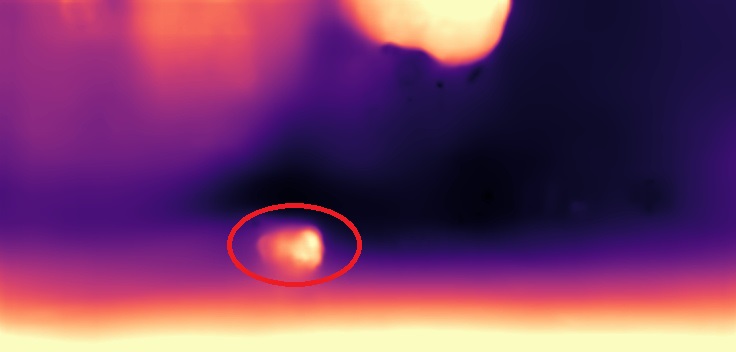}\vspace{4pt}
\includegraphics[width=1\linewidth]{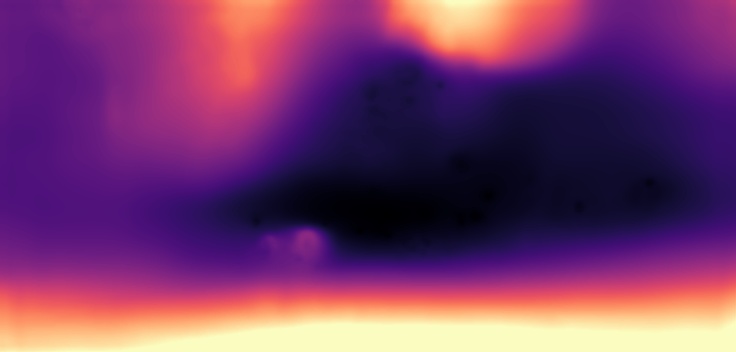}
\end{minipage}
\caption{The depth predictions from our model and baselines in driving at night.}
\label{fig: depth night}
\end{figure*}

\begin{figure}[htp]
    \centering
    \includegraphics[width=9.2cm]{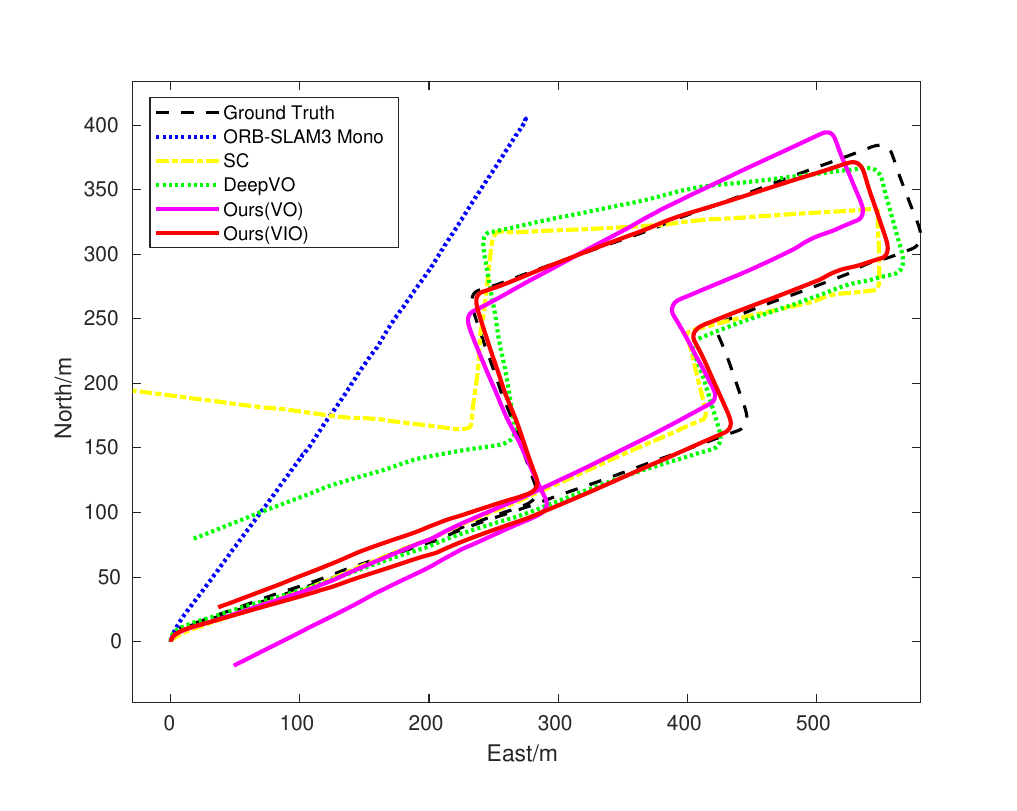}
    \caption{The generated trajectories in the night car-driving scenes}
    \label{fig: trajectory night}
\end{figure}

\subsection{Pose and Depth Evaluation on the Night Scenes of MVSEC Dataset}
To assess the effectiveness of our proposed model in extremely challenging night scenes, which typically feature low-light conditions and limited features, we conducted experiments on the MVSEC dataset. 
We compared our results against several other state-of-the-art methods, including ORB-SLAM3 \cite{ORB3}, a leading unsupervised-learning-based visual odometry (VO) method (SC-SfMLearner), and a supervised learning-based method (DeepVO).

The trajectories of each model on Evening3 are displayed in Figure \ref{fig: trajectory night}. It can be seen seen that the learning based VOs can robustly estimate vehicle ego-motion and generate trajectories in this challenging scene. However, the traditional model, ORB-SLAM3 Mono fails. This might be due to fact that ORB-SLAM3 Mono can extract and match enough feature points under low-light environment, and thus are not capable of performing effective feature tracking for pose estimation, leading to positioning failure. In contrast, deep learning networks excel at learning and extracting features 
and thus enable learning-based methods to extract sufficient and reliable visual feature for positioning. 

We use RMSE as the evaluation metric via Equation \ref{deqn_RMSE}, and further compare these quantitative results in Table \ref{tb: pose night},
\begin{equation}
\label{deqn_RMSE}
\text{Error} = \sqrt{\frac{1}{n}(\mathbf{x}_i-\mathbf{x}^\text{gt}_{i}) } \end{equation}
Where $\mathbf{x}_i$ represents the estimated three-dimensional location in $i th$ frame, and $\mathbf{x}^\text{gt}_{i}$ is the corresponding ground-truth location. The quantitative results show that the performance of our proposed VO model exceeds both unsupervised learning based VO, i.e. SC, and supervised learning based VO, i.e. DeepVO. It demonstrates that our framework can robustly and accurately estimate pose in low light environment.

The trajectories generated by each model on the Evening3 dataset are presented in Figure \ref{fig: trajectory night}. It is evident that the learning-based VOs are capable of accurately estimating vehicle ego-motion and producing trajectories in this challenging low-light and featureless scene. In contrast, the traditional model, ORB-SLAM3 Mono, fails to accurately estimate the pose in this scenario. As we know, in low-light conditions during nighttime, images often lack distinct features and textures. Traditional approaches to visual odometry encounter difficulties due to the inability of the front-end local feature extraction algorithm to capture essential elements such as corners and edges, which constitute the basis of features and textures. Consequently, effective feature tracking becomes problematic, leading to eventual position failure. In contrast, learning-based ego motion estimation models leverage deep neural networks to abstract global features from images, which are inherently more robust than local features. As training iterations progress, the network continuously enhances its ability to extract visual features, which is not feasible with hand-crafted feature-based visual methods. Therefore, learning-based ego motion estimation models exhibit superior performance in challenging environments.

In addition, we qualitatively evaluate the depth estimation performance of our framework in the challenging night scene, comparing with SC. 
Figure \ref{fig: depth night} shows input RGB image, depth maps from LIDAR data, depth estimation from SC, and depth estimation from our our model. 
Clearly, SC does not perform well when it is in the area of strong light. SC will misleadingly estimates depth in a shorter distance in strong-light condition. 
It might be because our model introduces a coarse-to-fine stage into depth estimation that enables DepthNet to overcome the influence of light condition, e.g. too strong or too dark.  

\subsection{Ablation Study}
The results of ablation study are presented in Table \ref{tb: ablation}. Firstly, to evaluate the impact of the attention module on pose estimation, we removed the attention mask and directly fed the fusion features of vision and inertial data into the subsequent LSTM and FC layers. We observed that the attention module can significantly improve the performance of PoseNet, particularly in terms of rotation estimation. Moreover, the inclusion of IMU data in the framework enhances the network's performance. This highlights the effectiveness of multimodal fusion, as the inertial data significantly contributes to rotation estimation, and the fusion of features can compensate for the limitations of individual modalities.

\begin{table}[h!]
\centering
\caption{A comparison of the scale factor estimation with or without our introduce scale recovery method.}
\renewcommand\arraystretch{1.5}
\begin{tabular}{cccc}
\hline
Method          & Scale      & $\mu$     & $\delta$ \\ \hline
Ours1  & x & 35.433 & 4.204 \\
Ours2  & \checkmark & 1.138  & 0.121 \\ \hline
\end{tabular}
\label{tb: scale factor}
\end{table}

Additionally, we evaluated the effectiveness of the LSTM network by removing the LSTM module and directly feeding the fused features into the FC layers. We found that adding the LSTM module improves the model's performance, which indicates the importance of incorporating temporal information into the model. Overall, our ablation study demonstrated the effectiveness of the attention module, the inclusion of IMU data, and the LSTM module in our framework, which all contribute to the accurate pose and depth estimation.

\begin{table}[h!]
\centering
\caption{The ablation study into framework modules}
\setlength{\tabcolsep}{1mm}{
\renewcommand{\arraystretch}{1.5}
\begin{tabular}{cccccccccc}
\hline
\multicolumn{1}{c}{\multirow{2}{*}{Method}} & \multirow{2}{*}{IMU} & \multirow{2}{*}{LSTM} & \multirow{2}{*}{Attention} & \multicolumn{2}{c}{Seq.09} & \multicolumn{2}{c}{Seq.10} & \multicolumn{2}{c}{Avg}                                    \\ \cline{5-10} 
\multicolumn{1}{c}{}                        &                      &                       &                            & $t_{rel}$     & $r_{rel}$     & $t_{rel}$     & $r_{rel}$     & \multicolumn{1}{c}{$t_{rel}$} & \multicolumn{1}{c}{$r_{rel}$} \\ \hline
Ours &   & $\checkmark$   & $\checkmark$                           & 8.84         & 1.31        & 7.34         & 1.98        & 8.09                         & 1.65                        \\ 
Ours                                        & $\checkmark$                     &                       & $\checkmark$                           & 14.57        & 0.56        & 10.92        & 1.73        & 12.75                        & 1.15                        \\ 
Ours                                        & $\checkmark$                     & $\checkmark$                      &                            & \textbf{5.06}         & 0.29        & 6.33         & 0.57        & 5.70                         & 0.43    \\ 
Ours & $\checkmark$ & $\checkmark$ & $\checkmark$ & 5.48         & \textbf{0.19}    & \textbf{5.37} & \textbf{0.43} & \textbf{5.43} & \textbf{0.31} \\ \hline
\end{tabular}}
\label{tb: ablation}
\end{table}

\begin{table}[h!]
\centering
\caption{The ablation study into sequence length}
\setlength{\tabcolsep}{1.7mm}{
\renewcommand{\arraystretch}{1.5}
\begin{tabular}{cccccccc}
\hline
\multicolumn{1}{c}{\multirow{2}{*}{Method}} & \multirow{2}{*}{Window length} & \multicolumn{2}{c}{Seq.09} & \multicolumn{2}{c}{Seq.10} & \multicolumn{2}{c}{Avg} \\ \cline{3-8}  &  & $t_\text{rel}$  & $r_\text{rel}$ & $t_\text{rel}$ & $r_\text{rel}$ & $t_\text{rel}$ & $r_\text{rel}$   \\ \hline
Ours1    &3        & 6.42         & 0.45        & \textbf{4.90}         & 0.55        & 5.66        & 0.5       \\ 
Ours2    &5 & \textbf{5.48}         & \textbf{0.19} & 5.37         & \textbf{0.43}         & \textbf{5.43}  & \textbf{0.31}       \\ \hline
\end{tabular}}
\label{tb: sequence ablation}
\end{table}

\begin{table*}[h!]
\small
\centering
\caption{The ablation into supervision signal and learning rate, "C" indicates Coarse Scale Recovery, "F" indicates Coarse-to-fine Scale Recovery, "D" indicates scaled depth predictions, "-" means this stage does not exist, "LR" indicates learning rate.}
\renewcommand\arraystretch{1.3} 
\begin{tabular}{ccccccccccc}
\hline
\multicolumn{1}{c}{\multirow{2}{*}{Method}}&\multicolumn{2}{c}{Supervision} &\multicolumn{2}{c}{LR} & \multicolumn{2}{c}{Seq.09} & \multicolumn{2}{c}{Seq.10} & \multicolumn{2}{c}{Avg} \\ \cline{2-3}   \cline{4-5}  \cline{6-11} 
        &C &F   &C  &F  & $t_\text{rel}$     & $r_\text{rel}$     & $t_\text{rel}$     & $r_\text{rel}$     & $t_\text{rel}$    & $r_\text{rel}$   \\ \hline
Ours1   &D  &D    &1e-4  &1e-5  & 8.01         & 1.07        & 9.06         & 1.45        & 8.54        & 1.26       \\ 
Ours2   & -  & x    &$\times$  &1e-4          & \textbf{5.46}         & 0.74        & 7.25         & 1.06         & 6.36       & 0.90       \\
Ours3   &D  & x  &1e-4 & 1e-5  & 5.48         & \textbf{0.19}        & \textbf{5.37}         & \textbf{0.43}        & \textbf{5.43}        & \textbf{0.31}       \\ 
\hline
\end{tabular}
\label{tb: supervision ablation}
\end{table*}

In order to investigate the impact of sequence length on pose estimation, we conducted an ablation study and compared the results with different sequence lengths. The comparison results are shown in Table \ref{tb: sequence ablation}.
Compared to a short sequence length of 3 frames, we observed that using a longer sequence window of 5 frames can improve both translation and rotation estimation accuracy. The improvement in rotation estimation is particularly significant. This indicates that using a longer sequence can provide more temporal context for the network to learn from and improve the accuracy of pose estimation.
It is worth noting that with an even longer sequence of 7 frames, we observed a slight decrease in performance. This could be attributed to the fact that a longer sequence can introduce more complex motion patterns and may make the optimization of network more challenging. Nonetheless, our results suggest that the selection of sequence length should be based on a trade-off between accuracy and complexity.

We conducted an investigation into the influence of adding scaled depth predictions and learned depth on pose estimation. The results are presented in Table \ref{tb: supervision ablation}. To verify the effectiveness of the learnable DepthNet on pose estimation, Ours1 was trained using coarse-to-fine scale recovery, but it utilized scaled depth predictions rather than the learned depths from DepthNet to form the total loss for bi-directional scale recovery. The learning rate used in this experiment was the same as that in Ours3. Although scaled depth predictions may have higher accuracy than learned depths, they are unlearnable and thus cannot be optimized through training iterations. As such, the potential for scaled depth predictions to improve pose estimation accuracy is limited.
On the other hand, to assess the effectiveness of scaled depth predictions on pose estimation, Ours2 was trained without incorporating scaled depth predictions into the framework, and instead directly utilized learned depth to form the total loss for bi-directional scale recovery. The learning rate used in this experiment was 1e-4. Since Ours2 lacks a coarse scale recovery stage, its pose estimates do not possess an absolute scale metric. Thus, the pose ground truth was used to scale Ours2's translation estimates. We believe that scaled depth predictions can not only recover coarse-scale pose but also improve pose estimation accuracy.
It is worth noting that Ours3, trained using our proposed strategy, outperforms Ours1 and Ours2.

\section{Conclusion}
 {Model-based monocular vision methods often struggle to perform well in challenging environments such as texture-less and low-lighting conditions, and they cannot provide accurate absolute-scale ego motion and depth estimates. To address these limitations, this study proposes a novel self-supervised learning framework for depth and egomotion estimation, which leverages learning-based pose and depth estimation modules. By harnessing the powerful feature extraction capabilities of deep neural networks, our framework is able to achieve robust performance even in complex lighting environments. Furthermore, our framework employs a coarse-to-fine scale recovery strategy to accurately and robustly estimate the pose and depth with absolute metric scale.}
The proposed approach utilizes a two-stage process to recover the scale of the depth predictions: First, the scaled depth predictions are used to supervise the PoseNet to recover the coarse scale pose. Second, a bi-directional scale recovery loss is employed to refine the performance of the pose network using learnable depth predictions.
Extensive experiments were conducted on two public datasets, namely KITTI RAW and MVSEC, to evaluate the effectiveness of the proposed framework. The results demonstrate that the proposed framework outperforms representative models and learning-based visual odometry (VO) and visual-inertial odometry (VIO) methods, achieving state-of-the-art performance. Furthermore, an ablation study was performed to analyze the effectiveness of each module in the proposed framework.

However, the study highlights that the learned scale may be limited to certain datasets and could be challenging to generalize into new domains. To address this limitation, future work will explore domain adaptation techniques for self-supervised egomotion and depth learning, enabling the trained model to quickly and efficiently adapt to new domains. Additionally, the study proposes to leverage inertial measurement unit (IMU) data for scale recovery and pose estimation to further improve the performance of the framework.  {Furthermore, our framework employs a relatively large deep neural network, i.e. ResNet, to construct feature extractors for PoseNet and DepthNet, which limits its real-time performance on mobile devices. To address this, in our future work, we plan to use knowledge distillation technology to develop a lightweight network structure. Additionally, despite the robustness of our framework in complex lighting environments, it has inherent limitations in adapting to different environments. To improve the generalization performance of our framework, we will explore domain adaptation techniques in our future research.}


\bibliographystyle{IEEEtran}
\bibliography{ref}

\par\noindent 
\parbox[t]{\linewidth}{
\noindent\parpic{\includegraphics[height=1.5in,width=1in,clip,keepaspectratio]{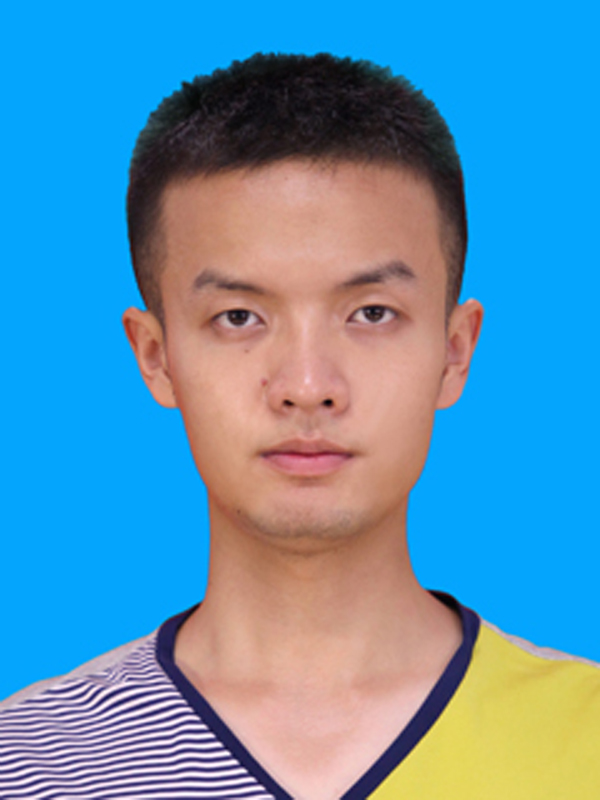}}
\noindent {\bf Hao Qu} is a Ph.D. student at College of Intelligence Science and Technology, National University of Defense Technology (China). Before that, he obtained his M.Eng. degree at National University of Defense Technology (China), and B.Eng.degree at National University of Defense Technology (China). His research interest lies in robotics, computer vision and navigation systems.
}
\vspace{2\baselineskip}

\par\noindent 
\parbox[t]{\linewidth}{
\noindent\parpic{\includegraphics[height=1.5in,width=1in,clip,keepaspectratio]{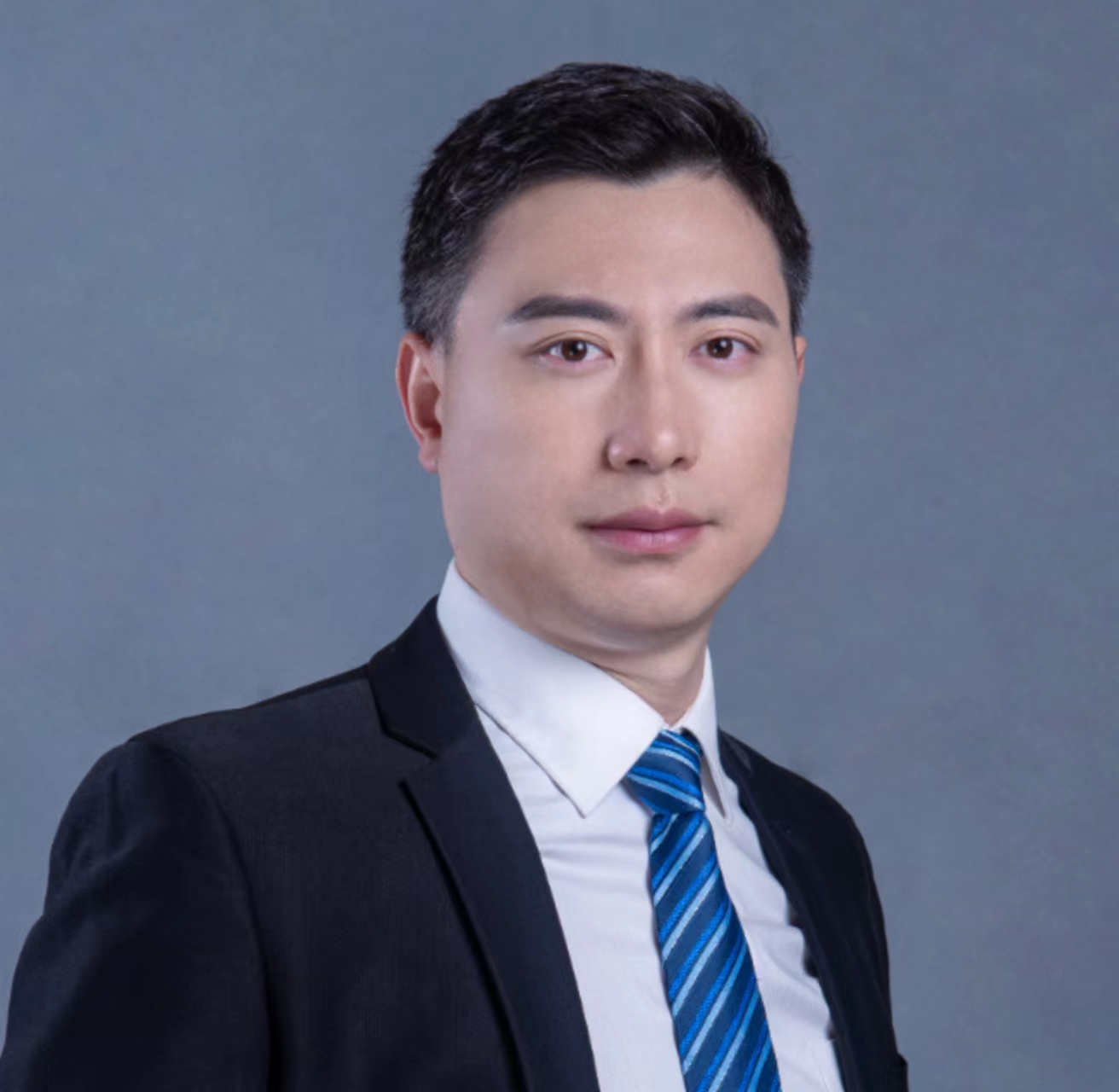}}
\noindent {\bf Lilian Zhang} received the B.S. degree from the School of Mechatronics and Automation,  National University of Defense Technology, China, in 2007, and the Ph.D. degree from  the Institute of Computer Science, University of Kiel, Germany, in 2013. He is currently an Associate Professor with the National University of Defense Technology.  His current  research interests include robotic vision and bionic navigation. 
}
\vspace{2\baselineskip}

\par\noindent 
\parbox[t]{\linewidth}{
\noindent\parpic{\includegraphics[height=1.5in,width=1in,clip,keepaspectratio]{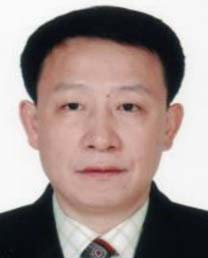}}
\noindent {\bf Xiaoping Hu} (Senior Member, IEEE) received the B.S. and M.S. degrees in automatic control systems and aircraft design from the College of Mechatronic Engineering and Automation, National University of Defense Technology, Changsha, China, in 1982 and 1985, respectively. He is currently a Professor with the National University of Defense Technology. His scientific interests include navigation, aircraft guidance, and control.
}
\vspace{2\baselineskip}

\par\noindent 
\parbox[t]{\linewidth}{
\noindent\parpic{\includegraphics[height=1.5in,width=1in,clip,keepaspectratio]{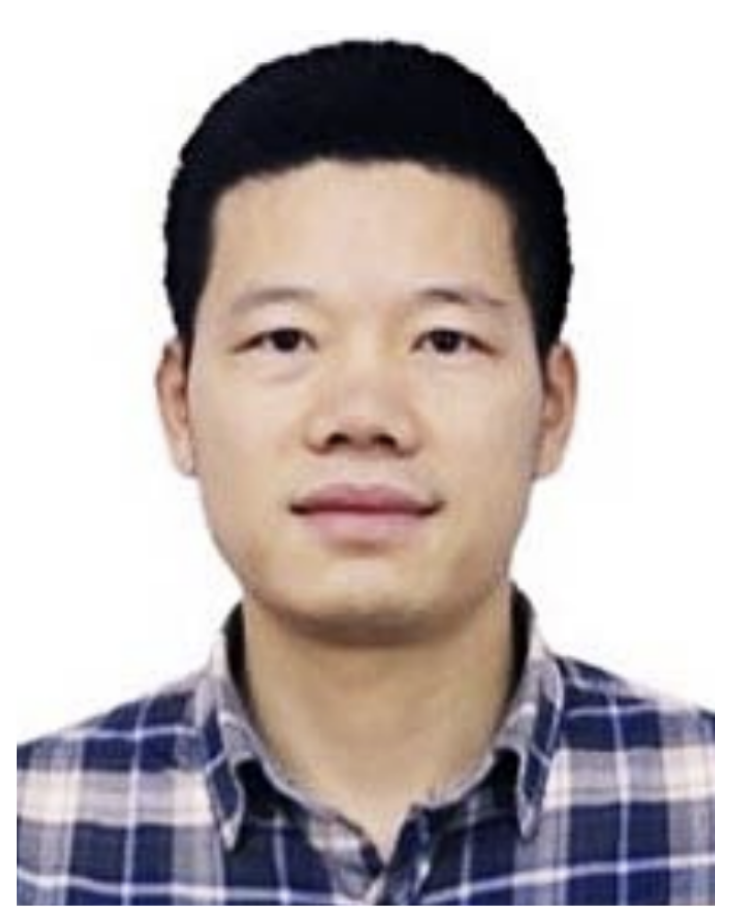}}
\noindent {\bf Xiaofeng He} received the B.S. and Ph.D. degrees from the School of Mechatronics and Automation, National University of Defense Technology, China, in 2001 and 2009, respectively.Since 2009, he has been an Associate Professor with the National University of Defense Technology. His current research interests include satellite navigation and deeply integrated navigation systems.
}
\vspace{2\baselineskip}

\par\noindent 
\parbox[t]{\linewidth}{
\noindent\parpic{\includegraphics[height=1.5in,width=1in,clip,keepaspectratio]{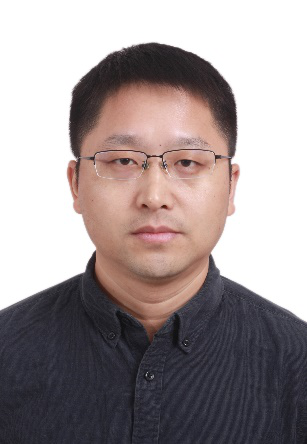}}
\noindent {\bf Xianfei Pan} received the Ph.D. degree in control science and engineering from the National University of Defense Technology, Changsha, China, in 2008. Currently, he is a vice professor of the College of Intelligence Science and Technology at the National University of Defense Technology. His current research interests include GPS/INS-integrated navigation system and indoor navigation system.
}
\vspace{2\baselineskip}

\par\noindent 
\parbox[t]{\linewidth}{
\noindent\parpic{\includegraphics[height=1.5in,width=1in,clip,keepaspectratio]{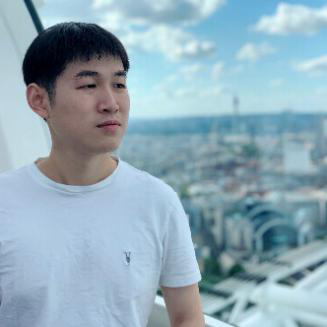}}
\noindent {\bf Changhao Chen} is a Lecturer at College of Intelligence Science and Technology, National University of Defense Technology. Before that, he obtained his Ph.D. degree at University of Oxford (UK), M.Eng. degree at National University of Defense Technology (China), and B.Eng. degree at Tongji University (China). His research interest lies in robotics, computer vision and cyberphysical systems.
}
\vspace{2\baselineskip}

\end{document}